\pdfoutput=1

\documentclass[11pt]{article}

\usepackage[]{ACL2023}

\usepackage{times}
\usepackage{latexsym}
\usepackage{amsfonts}

\usepackage[T1]{fontenc}

\usepackage[utf8]{inputenc}

\usepackage{microtype}

\usepackage{inconsolata}

\usepackage{multirow}
\usepackage{booktabs}
\usepackage{arydshln}
\usepackage{amsmath}
\usepackage{graphicx}
\usepackage{float}

\newcounter{HWNumberOfComments}
\stepcounter{HWNumberOfComments}

\newcounter{JSNumberOfComments}
\stepcounter{JSNumberOfComments}

\newcounter{RANumberOfComments}
\stepcounter{RANumberOfComments}

\title{Benchmarking zero-shot stance detection with FlanT5-XXL: Insights from training data, prompting, and decoding strategies into its near-SoTA performance
}
\author{Rachith Aiyappa \and Shruthi Senthilmani \and Jisun An \and Haewoon Kwak \and Yong-Yeol Ahn \\ 
        $\{$racball, shrusent, jisunan, hwkwak, yyahn$\}$@iu.edu \\
        Center for Complex Networks and  Systems \\
        Indiana University Bloomington}

\begin{document}
\maketitle
\begin{abstract}
We investigate the performance of LLM-based zero-shot stance detection on tweets.
Using FlanT5-XXL, an instruction-tuned open-source LLM, with the SemEval 2016 Tasks 6A, 6B, and P-Stance datasets, we study the performance and its variations under different prompts and decoding strategies, as well as the potential biases of the model. 
We show that the zero-shot approach can match or outperform state-of-the-art benchmarks, including fine-tuned models. 
We provide various insights into its performance including the sensitivity to instructions and prompts, the decoding strategies, the perplexity of the prompts, and to negations and oppositions present in prompts.  
Finally, we 
ensure that the LLM has not been trained on test datasets, and identify a positivity bias which may partially explain the performance differences across decoding strategies. 
\end{abstract}

\section{Introduction}

Stance detection is a fundamental computational task that is widely used across many disciplines such as political science and communication studies~\citep{wang2019survey,kuccuk2020stance}
Its goal is to extract the standpoint or \textit{stance} (e.g., Favor, Against, or Neutral) towards a target from a given text. 
Given that modern democratic societies make societal decisions by aggregating people's explicit stances through voting, estimation of peoples' stances is a useful task. 
While a representative survey is the gold standard, it falls short in scalability and cost~\citep{salganik2019bit}.   
Surveys can also produce biased results due to the people's tendency to report more socially acceptable positions even in anonymized surveys~\citep{krumpal2013determinants}.
It is also increasingly difficult to obtain a representative sample through phone call-based surveys as the communication channels shift away from telephones. 
As social media increasingly becomes the dominant platform and channel for self-expression and social communication~\citep{auxier2021social,vogels2022teens}, using it as social sensors~\citep{wang2019age}  for studying people's stances through their digital footprints (e.g., social media posts) is also becoming increasingly important~\citep{aldayel2021stance}.
However, inferring the stance of individuals using social media texts is challenging; they tend to be short and incoherent. They also tend to have more abbreviated writing, spelling errors as well as other information such as hyperlinks, hashtags, and usernames.

Since the rise of transformers, large language models (LLMs) built on the transformer architecture are now state-of-the-art natural language processing paradigms, with impressive performances across a huge range of tasks such as question answering and language translation. 
The most common approach to leverage pre-trained LLMs has been to use labeled data of a downstream task to fine-tune encoder models (e.g., BERT) to a specific task~\citep{devlin2018bert, kawintiranon2021knowledge, schiller2021stance}. 
Such fine-tuning approaches can benefit from both the general language understanding from the pre-training as well as the problem-specific thing, even without spending a huge amount of computing resources~\citep{wang2022pre}. 

More recently, the GPT family of models~\citep{radford2019language,brown2020language} birthed another powerful and even simpler paradigm of \textit{in-context learning} (``few-shot'' or ``zero-shot''). 
Instead of tuning any parameters of the model, it simply uses the input to guide the model to produce the desired output for downstream tasks. 
For instance, a few examples related to the task can be fed as the \textit{context} to the LLM. 

\begin{quote}
\hspace*{-0.5cm}
Tweet: Get a booster shot to be safe!\\
\hspace*{-0.5cm}
The stance towards vaccines is positive.

\hspace*{-0.5cm}
Tweet: The pandemic claimed many lives.\\
\hspace*{-0.5cm}
The stance towards the virus is negative.

\hspace*{-0.5cm}
Tweet: Jordan is the greatest of all time.\\
\hspace*{-0.5cm}
The stance towards Michael Jordan is \rule{0.8cm}{0.15mm}    
\end{quote}
In-context learning requires very few, if any labeled data, and the accuracy of labeling is not a major concern~\citep{min2022rethinking}. 
This makes it particularly well-suited for cases with limited annotated data and minimal computing resources. 
We can even remove the examples and directly ask the question
This is called zero-shot inference and it works surprisingly well for many standard NLP tasks~\cite{radford2019language}. 
In the context of stance detection, this raises the following questions, 
(1) How capable (and sensitive) are LLMs for stance detection in a truly zero-shot setting?
(2) How sensitive is the performance to factors like instructions, prompts, model biases, and decoding strategies? 

With the growing trove of LLMs, and the increasing usage of in-context learning with the rise of chat LLMs, it is important to quantify and understand the performance of state-of-the-art LLMs on stance detection, which is not a common task that LLMs are evaluated on.
%
\section{Related work}
The past decade has seen an increased interest in the stance detection task, especially since the SemEval 2016 Task 6 stance detection competition~\citep{mohammad2016dataset}. For the latest review of stance detection models, we refer the readers to~\citet{alturayeif2023systematic}. 

Traditional models used human-engineered linguistic features such as n-grams and lexicon-based sentiment, in combination with classifiers like support vector machine (SVMs)~\citep{hacohen2017stance,dey2017twitter,mohammad2017stance},  (Bi-)LSTMs~\citep{augenstein2016stance}, and RNNs~\citep{mohammad2016semeval} to extract stance from text. 
This approach required significant feature engineering and large training data.

With the advent of transformer architecture, fine-tuning pre-trained LLMs like BERT and RoBERTa showed impressive performance~\citep{ghosh2019stance,zhao2020pretrained,schiller2021stance,he2022infusing,alturayeif2023systematic}
Because these approaches still rely on the availability of substantial training data, the few-shot~\citep{brown2020language} and zero-shot~\citep{wei2021finetuned} capabilities of LLMs have garnered much interest.
Recent work tested LLMs in a ``cross-target stance detection'' setting where the LLM was fine-tuned on stance towards certain topics but evaluated on topics not seen during training~\citep{allaway2020zero,allaway-etal-2021-adversarial,gambini2022tweets2stance}. 
Although this a more challenging task, it still relies on the availability of labeled data for a few targets to fine-tune existing LLMs and ``teach'' it the task, or create new LLMs, or on the availability of external knowledge bases~\cite{liu2021enhancing,luo2022exploiting}

Only a handful of recent studies investigate the zero-shot stance detection performance of LLMs~\cite{ liang2022jointcl,wen2023zero}, including studies of ChatGPT's---an instruction-tuned GPT model---performance on English datasets~\citep{zhang2022would,kocon2023chatgpt,zhang2023investigating}, that of open-source LLMs like BERT, BART, and RoBERTa on non-English datasets~\citep{gambini2022tweets2stance,zhao2023c}, and are in cross-linguistic settings~\citep{vamvas2020x}.
The most relevant recent work by \citet{cruickshank2023use} who test the performance of some of the latest open-source LLMs (Flan-Alpaca-GPT4-T5, Flan-UL2). 
However, these studies used weaker baselines and did not provide much insight into the grounded performance measures and other dependence on various factors that we study here. 
For instance, there is a potential that the model(s), may have seen the task data during training~\cite{aiyappa-etal-2023-trust,li2023task}, which most existing studies overlooked.
Similarly, an in-depth analysis of the model(s) biases and its sensitivity to decoding strategies during inference is missing, although~\citet{cruickshank2023use} offer some insights into the effect of different prompting schemes (zero-shot, few-shot, few-shot with reasoning).

Despite the growing popularity of stance detection, to the best of our knowledge, an in-depth analysis of the off-the-shelf, zero-shot, performance of an open-source generative LLM on English tweets is still missing.  
Our study aims to fill this knowledge gap. 
\section{Dataset}
We employ two popular stance detection datasets (1) SemEval 2016 Task6~\citep{mohammad2016dataset}, and (2) P-Stance~\citep{li2021p}.
As an initial step, we limit ourselves to English language datasets originating from the United States Twitter discourse. 
This choice is motivated by (1) a majority of LLMs specializing in English which can help us isolate the robustness, variation, and sensitivity of the model, (2) the challenging nature of tweets---presence of hashtags, abbreviations, URLs, spelling mistakes, and limited character count, and (3) the US population being the most active demographic on Twitter.\footnote{\url{https://bit.ly/3QZLRDO}} 
Additionally, we choose the simplest flavor of stance detection where each tweet has a stance towards a single pre-specified target.
We leave it to future work to explore more complex settings like non-English tweets, or when the target is not pre-specified, or when each tweet has stances towards multiple targets~\citep{aldayel2021stance}, or when the context is longer text than tweets.
\subsection{SemEval 2016 Task 6A and B}
This is the most popular dataset for stance detection on tweets~\citep{mohammad2016dataset}, where 
the task is to infer a stance---Favor, Against, or Neutral/None---of a tweet towards a pre-selected target. 
Task 6A includes targets with annotated training data, which are Hillary Clinton (HC), Feminist Movement (FM), Legalization of Abortion (LA), Atheism (AT), and Climate Change is a Real Concern (CC).
Task 6B on the other hand, includes a single target, Donald Trump (DT), for which no labeled data is available. 
The tweets from both tasks that were not included for stance annotation were released as a part of the ``domain corpus'' (Appendix~\ref{app:task6} Tab.~\ref{tab:domain_corpus}).
Since we are testing the LLM's performance in a zero-shot setting, we do not make use of any of the training data. 
Task 6A and 6B have 1249 and 707 tweets in their test sets respectively (Appendix~\ref{app:task6} Tabs.~\ref{table:Task6A_data}  and~\ref{table:Task6B_data}). 
\subsection{P-Stance}
This dataset was created in a similar spirit to SemEval 2016 Task 6 while being approximately 5 times larger~\citep{li2021p}.
It contains tweets about three pre-defined targets and the goal is to infer a binary stance---Favor or Against---of a tweet toward targets (``Donald Trump,'' ``Joe Biden,'' ``Bernie Sanders''). 
Appendix~\ref{app:pstance} Tab.~\ref{tab:pstance_data} shows the number of tweets in the training, validation, and test datasets. 
There are 796, 745, and 632 tweets in the test sets of Trump, Biden, and Sanders, respectively.
\section{Model}
For our experiments, we use FlanT5-XXL~\citep{chung2022scaling}, Google's 11-billion parameter, instruction-tuned, encoder-decoder language model based on T5~\citep{raffel2020exploring}.
The instruction-tuning was done in a zero/few-shot and chain of thought setting, on several NLP datasets\footnote{SemEval 2016 Tasks 6A, 6B, and P-Stance (\S\ref{sec:instruction-tuning}) were not included.} using a range of instruction tuning methods like input inversion, and template generation~\citep{flan,wei2021finetuned}. 
We choose this model because (1) it is documented that instruction-tuned LLMs, 
as \textit{small} as 1B parameters are often better than their non-instruction-tuned counterparts with many more parameters~\citep{wei2021finetuned,ouyang2022training} (see Appendix~\ref{app:preliminary}), (2)~\citet{wang2022language} showed that instruction-tuned encoder-decoder models outperform the decoder-only counterparts in various NLP tasks in the zero-shot setting, (3) closed-source LLMs like ChatGPT pose concerns about reproducibility and test data contamination~\cite{aiyappa-etal-2023-trust}, and 
(4) 
FlanT5-XXL's excellent performance across many tasks~\cite{chia2023instructeval}.
We use FlanT5-XXL and its tokenizer available from Hugging Face.\footnote{https://huggingface.co/google/flan-t5-xxl}
The model requires around 40GB of GPU RAM to load and use.

\section{Methodology}

Although the task of stance detection is typically considered a classification task, we reframe it as a \emph{question-answering} task, to utilize FlanT5-XXL's generative nature~\citep{schick2020exploiting}.

We use the following \textit{prompt} template,
\begin{quote}
$\langle$Instruction$\rangle$ \\
Statement: $\langle$Tweet$\rangle$ \\
Question: $\langle$Question$\rangle$
\end{quote}
$\langle$Instruction$\rangle$ is a placeholder to define the output requirement. 
For example, ``Answer the question with positive, negative, or neutral.'' 
$\langle$Tweet$\rangle$ contains the tweet text. 
$\langle$Question$\rangle$ guides the LLM to extract the stance from the statement. 
In the following sections, we explain how we design the instructions (\S5.1), prompts (\S5.2), and decoding strategies to select the output (\S5.3). 
This enables us to measure FlanT5-XXL's sensitivity to these design choices. 
We then introduce metrics to evaluate FlanT5-XXL's performance (\S5.4) and baseline models (\S5.5) to compare it against. 


%
\subsection{Prompt Design}
It is well known that the performance of LLMs on any task is sensitive to the prompt~\citep{zhao2021calibrate}. 
Acknowledging that there are many ways to design prompts, we propose prompts that are both natural to the task and, more importantly, may allow us to gain a few insights into the functioning of FlanT5-XXL (see Tab.~\ref{tab:prompts}). 

\begin{table}[h]
\small 
\centering
\begin{tabular}{|c|l|c|}
\hline
\multirow{2}{*}{\centering ID} & \multirow{2}{*}{Prompt Format} & \multirow{2}{*}{Out. Req.} \\
& & \\
\hline
\multirow{3}{*}{\centering 1} & $\langle$Instruction$\rangle$ & \multirow{4}{*}{\shortstack{positive/ \\ negative/ \\ neutral }} \\
& Statement: $\langle$tweet$\rangle$ & \\
& \begin{tabular}[t]{@{}l@{}} Question: What is the attitude \\ of the statement to $\langle$target$\rangle$?\end{tabular} & \\
\hline
\multirow{3}{*}{\centering 2} & $\langle$Instruction$\rangle$ & \multirow{4}{*}{\shortstack{favor/ \\ against/ \\ neutral}} \\
& Statement: $\langle$tweet$\rangle$ & \\
& \begin{tabular}[t]{@{}l@{}} Question: What is the stance \\ of the statement to $\langle$target$\rangle$?\end{tabular} & \\
\hline
\multirow{3}{*}{\centering 3f} & $\langle$Instruction$\rangle$ & \multirow{4}{*}{\shortstack{true/ \\ false/ \\ neutral}} \\
& Statement: $\langle$tweet$\rangle$ & \\
& \begin{tabular}[t]{@{}l@{}} Question: The statement \\ is in favor of $\langle$target$\rangle$.\end{tabular} & \\
\hline
\multirow{3}{*}{\centering 3a} & $\langle$Instruction$\rangle$ & \multirow{4}{*}{\shortstack{true/ \\ false/ \\ neutral}} \\
& Statement: $\langle$tweet$\rangle$ & \\
& \begin{tabular}[t]{@{}l@{}} Question: The statement \\ is against $\langle$target$\rangle$.\end{tabular} & \\
\hline
\multirow{3}{*}{\centering 4f\_not} & $\langle$Instruction$\rangle$ & \multirow{4}{*}{\shortstack{true/ \\ false/ \\ neutral}} \\
& Statement: $\langle$tweet$\rangle$ & \\
& \begin{tabular}[t]{@{}l@{}} Question: The statement \\ is not in favor of $\langle$target$\rangle$.\end{tabular} & \\
\hline
\multirow{3}{*}{\centering 4a\_not} & $\langle$Instruction$\rangle$ & \multirow{4}{*}{\shortstack{true/ \\ false/ \\ neutral}} \\
& Statement: $\langle$tweet$\rangle$ & \\
& \begin{tabular}[t]{@{}l@{}} Question: The statement \\ is not against $\langle$target$\rangle$.\end{tabular} & \\
\hline
\end{tabular}
\caption{Different types of prompts used to evaluate FlanT5-XXL's performance on SemEval 2016 Task6. $\langle$Instruction$\rangle$ are chosen from Appendix~\ref{app:instrutions}}
\label{tab:prompts}
\end{table}

For instance, Prompt 2 in Tab.~\ref{tab:prompts} directly asks the model to provide the intended stance while Prompts 3f and 3a pick a stance label (``favor'' or ``against'') and ask the model a true/false question. 
Given the symmetric nature of these prompts and that ``favor'' and ``against'' are antonyms from the perspective of the datasets, an LLM that answers true (false) to Prompt 3f, should ideally answer false (true) to Prompt 3a, unless the answer is ``neutral'' in which case both 3f and 3a should answer the same. 
In other words, ideally, the performance of both 3f and 3a should be exactly the same if the LLM is perfectly logical. 
In a similar spirit, responses to 3f (3a) and 4f\_not (4a\_not) should be opposites.
While more sophisticated prompt designing strategies are available~\citep{liu2021pre}, these may require access to annotated training data to be effective~\citep{zhou2022large,fernando2023promptbreeder}. 
We choose to restrict ourselves strictly to a zero-shot setting where we do not expose the model to any kind of task-specific data during test time or to design better prompts.
In other words, our experiments reveal a lower bound of FlanT5-XXL's zero-shot performance on the datasets and its optimal performance can be higher. 
\subsection{Instructions}
The nine $\langle$Instruction$\rangle$s for each Prompt ID are listed in Appendix~\ref{app:instrutions}. 
These instructions were generated using ChatGPT, by asking it to paraphrase a seed instruction ``Answer the question with $\langle$opt-1$\rangle$, $\langle$opt-2$\rangle$, or $\langle$opt-3$\rangle$.'' created by the authors. 
In principle, any generative language model can be used to paraphrase this seed instruction but we choose not to use FlanT5 or T5 to avoid a potential confounder~\citep{gonen2022demystifying}. 
\subsection{Output selection}~\label{sec:output_selection}
By default, FlanT5 uses \textit{greedy} decoding to answer the question posed to it. 
Given the context (prompt, $x$), it selects a token, $y_i$, with the highest probability (Eq.~\ref{eq:greedy}), adds this token to its context, proceeds until a stopping condition is met, and prints out all the new tokens as the output. 
In our experiments, the stopping condition is the default end-of-sentence token. 
\begin{equation}
    \arg\max_i \prod_{j=1}^{i}\log P(y_j \mid x, y_{1, \cdots, j-1})\label{eq:greedy}
\end{equation}
However, it is known that LLMs have biases that make them more likely to output certain tokens over others (e.g., tokens towards the end of
the prompt in Tab.~\ref{tab:prompts}) and other decoding methods such as ``affine transformation'' (AfT) have been proposed~\citep{zhao2021calibrate}.
Here, the key idea is to estimate the model’s bias towards certain answers by feeding in a context-free (cf) input and then use this result to compensate. 
In our case the context-free input is the one that does not contain the $\langle$tweet$\rangle$ in Tab.~\ref{tab:prompts}. 
We then apply a weight matrix $\textbf{W}$ and a bias vector $\textbf{b}$ to the original token probabilities vector $\hat{\textbf{p}}$ (obtained via greedy decoding) to get the new probabilities vector $\textbf{q}$ where
\begin{equation}
    \textbf{q} = softmax(\textbf{W}\hat{\textbf{p}} + \textbf{b})\label{eq:AfT}
\end{equation}
The strategy is then to pick the token with the largest $\textbf{q}$, i.e. $\text{argmax}(\textbf{q})$. 
Similar to~\citet{zhao2021calibrate} we set $\textbf{W} = \text{diag}(\textbf{p}_{\textbf{cf}})^{-1}$ where $\textbf{p}_{\textbf{cf}}$ is the token probabilities vector (obtained via greedy decoding) in a context-free setting, and $\textbf{b}$ to an all-zero vector. 

Another strategy to overcome this is to use pointwise mutual information (PMI)~\citep{holtzman2021surface}, where $\text{PMI}(x,y_i)$ estimates how much more likely is it for token $y_i$ to occur in the presence of context $x$, than by itself, i.e. $\frac{P(y_i|x)}{P(y_i)}$.
Now the strategy is to pick the token with the largest PMI value, i.e. 
\begin{equation}
\begin{split}
    \arg\max_i \frac{P(y_i|x)}{P(y_i|x_{cf})} 
    \approx \arg\max_i \frac{P(y_i|x)}{P(y_i|x_{cf})}.\label{eq:pmi}
\end{split}
\end{equation}
Since LLMs are built in a conditional setting, $P(y_i)$ is approximated as $P(y_i|x_{cf})$ which extracts the likelihood of $y_i$ occurring in a context-free setting. This makes Eq.~\ref{eq:pmi} similar to Eq.~\ref{eq:AfT} since for token $y_i$, $(W\hat{p})_i=\frac{\hat{p}_i}{(p_{cf})_i}=\frac{P(y_i|x)}{P(y_i|x_{cf})}$, and the only difference is that AfT strategy uses softmax while PMI uses max.

In our experiments, we compare the outputs obtained via `greedy,' `AfT,' and `PMI' decoding strategies. 
\subsection{Evaluation Metrics}
In line with the official competition metric of the SemEval 2016 Task 6 and subsequent works, we evaluate the performance of FlanT5-XXL using the macro average ($F_{avg}$)  of the F-scores on the `favor' and `against' classes (see Appendix~\ref{app:evaluation_metrics}). This can be seen as a micro average of F-scores across targets.
\subsection{Baseline Models}~\label{sec:baseline}
We test the zero-shot performance of FlanT5-XXL on SemEval 2016 Task 6A, 6B, and P-Stance.
Given that we use FlanT5-XXL off-the-shelf without any modifications to it, a fair comparison would be to pit it against LLMs utilized similarly. 
However, zero-shot stance detection is not a common task in which off-the-shelf LLMs have been rigorously tested.
Thus, as our baselines, we choose the best available models, identified by ~\citet{alturayeif2023systematic} and the authors.
These mostly comprise models that have been trained/fine-tuned for stance detection and cover a range of approaches---from fully/weakly supervised featured-based approaches to more recent fine-tuned LLMs.
We list the strongest baselines available in chronological order and summarize their approach in more detail in Appendix~\ref{app:baselines}. Additionally, we wish to highlight that ``state-of-the-art'' baselines used in previous studies often do not contain the actual state-of-the-art (see Appendix~\ref{app:questioning_baselines}). 

\paragraph{SemEval 2016 Task 6A}
(1) LibSVM (2017): This is a supervised model trained only on the task's training data~\citep{hacohen2017stance}.
(2) PE-HCN (2020)~\citep{zhao2020pretrained}: This fine-tunes RoBERTA on the task's training data and is currently state-of-the-art to the best of our knowledge. (3) TimeLM (2022): This is a BERT model that is periodically fine-tuned on the latest tweets collected in quarterly time intervals, and further fine-tuned on the task's training data~\citep{loureiro2022timelms}, and 
(4) ChatGPT (2023): \citet{kocon2023chatgpt} evaluated the performance of ChatGPT on zero-shot stance detection.


\paragraph{SemEval 2016 Task 6B}
(1) BiLSTM (2016): A weakly-supervised approach that uses the domain corpus~\citep{augenstein2016stance}, 
(2) SVM (2017): A two-stage classification system that first classifies tweets as neutral/non-neutral and then the non-neutral tweets are classified as positive vs. negative~\citep{dey2017twitter}, 
(3) POLITICS (2022): ~\citet{liu2022politics} fine-tuned RoBERTa on additional objectives on a newly curated dataset, and further on Task 6A's data, and 
(4) PT-HCL (2022): A fine-tuned BERT model on task 6A data~\citep{liang2022zero}.

\paragraph{P-Stance}
(1) RoBERTA KSCP (2022):~\citet{zheng2022knowledge} propose Knowledge Stimulated Contrastive Prompting which fine-tuned a RoBERTa model by including the target in the prompt.
However, they report a drop in performance in a cross-target setting---when the LLM is fine-tuned, say on `Trump' (`RoBERTa KSCP trump') and tested on a say, `Biden.' 
(2) RoBERTa KEPT (2023):~\citet{huang2023knowledge} propose Knowledge Stimulated Contrastive Prompting, which fine-tunes RoBERTa on the task's training data and additional data created by substituting the targets and stance labels with their synonyms. 
This work does not present scores for the target ``Bernie Sanders.'' 
(3) COT ChatGPT (2023):~\citet{zhang2023investigating} evaluates ChatGPT's performance in a Chain-of-Thought setting, and 
(4) Zero-Shot ChatGPT (2023):~\citep{zhang2023investigating} test the zero-shot capability of ChatGPT---the only baseline that offers a fair comparison to our work. 
This is also the current SoTA in a zero-shot setting.
\section{Results}~\label{sec:results}
\begin{figure*}[t]
    \centering
    \includegraphics[width=1\textwidth, trim={0 2.2in 0 2.2in}]{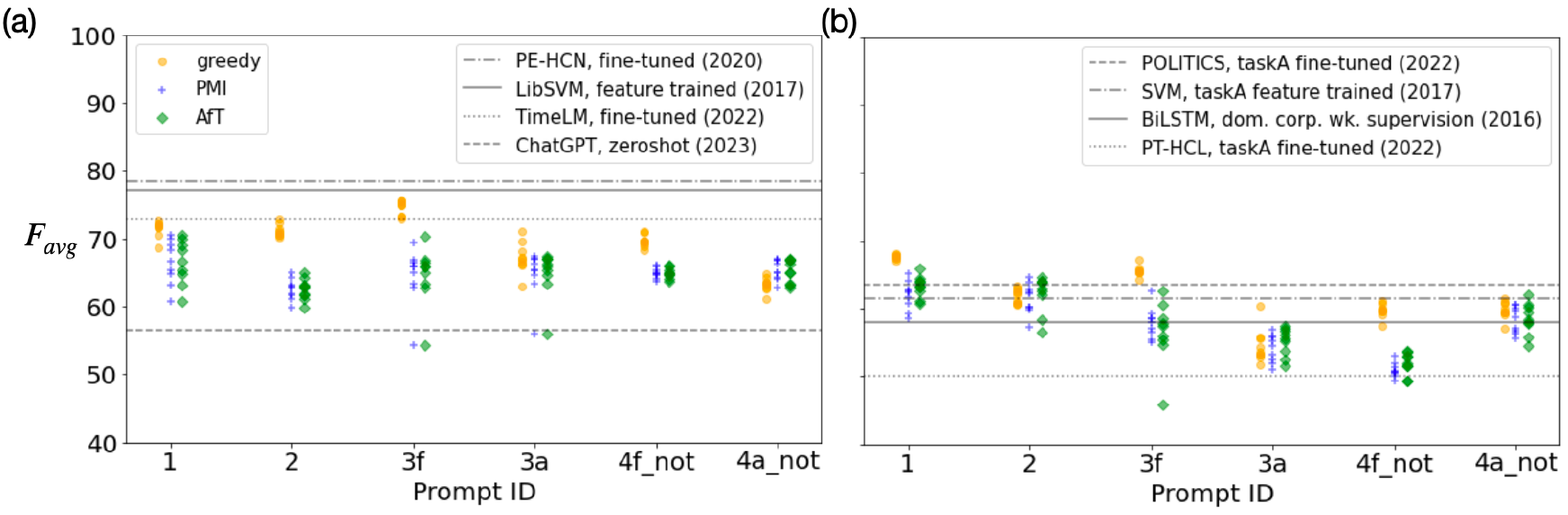}
    \caption{FlanT5-XXL is capable of outperforming state-of-the-art baselines of stance detection in a zero-shot setting and matching the performance of fine-tuned LLMs in SemEval 2016 Task 6A while beating it in Task 6B. The $F_{avg}$ scores of FlanT5-XXL on (a) Task 6A and (b) Task 6B of SemEval 2016 are shown in comparison against some of the best-performing models. Each label on the $x$-axis corresponds to a prompt (see Tab.~\ref{tab:prompts}) and each point on a given prompt ID corresponds to an instruction (see Appendix.~\ref{app:instrutions}). The results of three decoding strategies---greedy, PMI, and AfT---are also shown.}\label{fig:taskAB}
\end{figure*}

\begin{figure*}[t]
    \centering
    \includegraphics[width=1\textwidth, trim={0 2.8in 0 3in}]{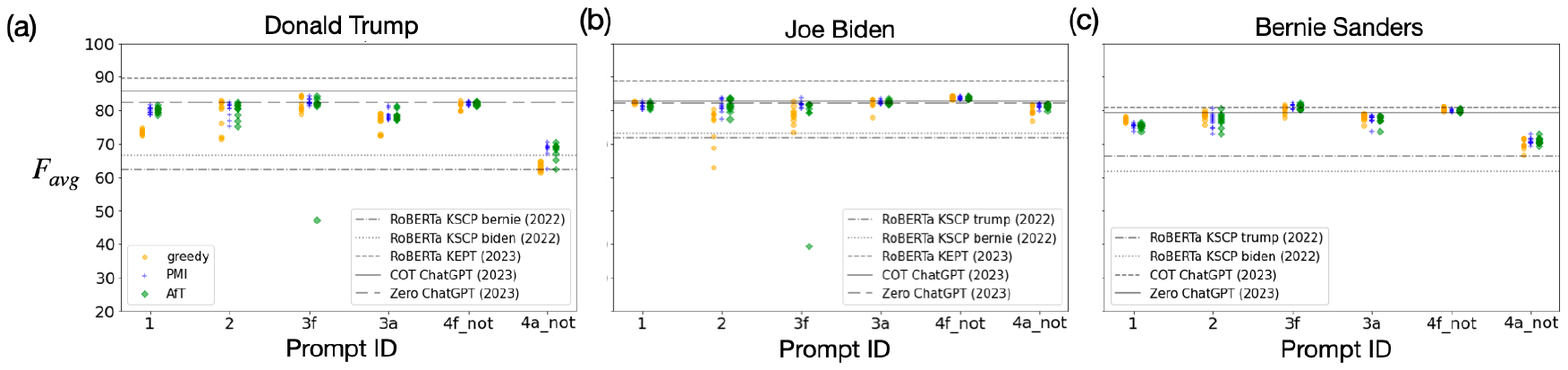}
    \caption{FlanT5-XXL is capable of outperforming state-of-the-art baselines in stance detection in a zero-shot setting in the P-Stance dataset. The $F_{avg}$ scores of FlanT5-XXL on different targets in P-Stance are shown in comparison against some of the best-performing models. Each label on the x-axis corresponds to a prompt (see Tab.~\ref{tab:prompts}) and each point on a given prompt ID corresponds to an instruction (see Appendix.~\ref{app:instrutions}). The results of two decoding strategies---greedy and PMI---are also shown.}\label{fig:pstance}
\end{figure*}

The performance of FlanT5-XXL on SemEval 2016 Task 6A, 6B, and P-Stance datasets evaluated using $F_{avg}$ is shown in Figs.~\ref{fig:taskAB} and~\ref{fig:pstance},  respectively. 
It also shows (1) the $F_{avg}$ scores of the baseline models, (2) the performance via greedy, PMI, and AfT decoding strategies, and (3) the variance (Appendix~\ref{app:instruction_sensitivity}) in performance across different prompt types (Tab.~\ref{tab:prompts}) and instructions (Appendix~\ref{app:instrutions}). 

\paragraph{SemEval 2016 Task 6} 
Fig.~\ref{fig:taskAB} 
shows that the performance of FlanT5-XXL on Task 6A and 6B.
An independent sample t-test showed that the difference in the mean $F_{avg}$ scores between prompt pairs is mostly statistically significant ($p<0.05$) when greedy decoding is used but such is not the case when PMI or AfT decoding is used (Appendix~\ref{app:statsig} Fig.~\ref{fig:statsig_semeval}).
However, in both tasks, we observe a significant difference between the mean $F_{avg}$ scores for prompts (1) 3f and 4f\_not, (2) 3a and 4a\_not, and (3) 3f and 3a, in the greedy decoding (independent sample t-test $p\ll0.05$).
This points to the difficulty of FlanT5-XXL, much like humans~\citep{wales1969so}, to answer questions framed as an opposition (3a: ``against") or with negation (4f\_not: ``not in favor of"; 4a\_not: ``not against"). 
We capture these differences in performance between prompts, by measuring the correlation between the perplexity (Appendix ~\ref{app:perp}) of the prompt and its performance, and find a significant negative correlation in multiple settings (Appendix ~\ref{app:perp}, Fig.~\ref{fig:perplexities}).

Similarly, we observe that the performance difference between greedy and PMI or AfT decoding on the same prompt are all statistically significant ($p<0.05$) except on prompts 3a on Task 6A and prompts 2, 3a, and 4a\_not in Task 6B but no significant differences between PMI and AfT decoding.
We also note that PMI or AfT decoding shows a significant drop in performance and an increase in standard mean error (see Appendix~\ref{app:instruction_sensitivity} Tab.~\ref{tab:sme_greed_pmi}) when compared to greedy.\footnote{The differences in standard mean errors are not significant ($p>0.05$)} This is surprising given that our expectation was for PMI or AfT to do better than greedy (\S~\ref{sec:output_selection}) and we posit the interplay between the model's positivity bias (Appendix ~\ref{app:positivitiy_bias}) and ``against'' being a majority class in the dataset (Appendix ~\ref{app:task6} Tab.~\ref{table:Task6A_data}, and Tab.~\ref{table:Task6B_data}) to be a reason for this (see Appendix~\ref{app:PMI}). Nevertheless, this indicates that using the model off-the-shelf, i.e. in the default greedy decoding, suffices in most cases. 
In contrast to prompt sensitivity, we find that wordings of the instructions have minimal effect on performance (see. Tab.~\ref{tab:sme_taskA}, and Tab.~\ref{tab:sme_taskB} in Appendix ~\ref{app:instruction_sensitivity}), as long as they are paraphrases. 

Overall, the performance of FlanT5-XXL is close to SoTA on Task 6A and even outperforms SoTA on Task 6B (Tabs.~\ref{tab:taskA} and~\ref{tab:taskB}). More impressively, its performance is much better than previous zero-shot attempts on task 6A (ChatGPT 2023) and task 6B, and more recent approaches which fine-tune LLMs in a semi-supervised (e.g., TimeLM, PT-HCL) or weakly-supervised (e.g., BiLSTM, PT-HCL) manner. Considering that FlanT5-XXL has never been exposed to an English Twitter stance detection task (Appendix~\ref{app:contamination}), combined with its off-the-shelf benefit, this result is particularly impressive. 

The best-performing prompts, in terms of the overall $F_{avg}$ score for Task 6A and 6B are shown in Appendix~\ref{app:semeval_performance} Tabs.~\ref{tab:taskA}  and~\ref{tab:taskB}, respectively along with the baseline models. This is prompt 3f with instruction 2 for Task 6A, and prompt 1 with instruction 6 for Task 6B. Tab.~\ref{tab:taskB} also shows that the performance of prompt 3f with instruction 2---the best performer on Task 6A (Tab.~\ref{tab:taskA})---is close (0.68 lesser) to the best performer on Task 6B, but still beats the previous state of the art. 
This suggests that prompt 3f with instruction 2 (i.e., ``The question needs to be answered with either true, false, or neutral'') can be considered the overall best performer on zero-shot stance detection on the SemEval 2016 Task 6.
The tables also show the target-specific $F_{avg}$ score, highlighting FlanT5-XXL's SoTA performance for the target ``Hillary Clinton.''  
The $F_{avg}$ scores, based on greedy decoding, per prompt, per instruction, and per target, for Task 6A is shown in Appendix~\ref{app:taskA}, and for Task 6B in Appendix~\ref{app:taskB}.

Lastly, we explore if pre-processing tweets as expanding abbreviations and splitting hashtags~\citep{ghosh2019stance} improves model performance and find this to be the case (Appendix ~\ref{app:preprocessing}).

\paragraph{P-Stance}

Fig.~\ref{fig:pstance} shows the performance of FlanT5-XXL on the P-Stance dataset. 
As observed in the SemEval 2016, greedy decoding often results in statistically significant differences between prompt pairs but the other decoding strategies produce more consistent results. 
We still observe certain difficulties for FlanT5-XXL to process opposition and negations but are less stark compared to SemEval 2016.  (Appendix~\ref{app:statsig}, Fig.~\ref{fig:statsig_pstance}).
Once again, we capture these differences in performance between prompts, by measuring the correlation between the perplexity (Appendix~\ref{app:perp}) of the prompt and its $F_{avg}$ score, and find a significant negative correlation in multiple settings (Appendix~\ref{app:perp}, Fig.~\ref{fig:perplexities_pstance}).

Unlike in SemEval 2016, we observe less of a difference in performance between greedy and PMI/AfT decoding strategies on the same prompt but like on SemEval, performances between PMI and AfT decoding strategies do not differ significantly on the same prompt (Fig.~\ref{fig:pstance}).\footnote{those which do differ significantly ($p<0.05$) are:  Donald Trump---prompts 1, 3a, and 4a\_not are between greedy and PMI/AfT. 
Joe Biden---prompts 1, 2, 4a\_not between greedy and PMI/AfT, while 3f is only different between greedy and PMI. 
Bernie Sanders---prompts 1, 3f, 4f\_not between greedy and PMI/AfT.}
We once again find that the wordings of instructions have a minimal effect on performance (see. Tab.~\ref{tab:sme_pstance} in Appendix ~\ref{app:instruction_sensitivity}), as long as they are paraphrases. 
Additionally, unlike in the SemEval 2016 Task 6, the standard mean errors in PMI/AfT decoding are often smaller than greedy decoding (Appendix~\ref{app:instruction_sensitivity} Tab.~\ref{tab:sme_pstance}), but the difference is not statistically significant.
Overall, in a majority of cases, greedy still outperforms other decoding strategies bolstering the ability of FlanT5-XXL to be used off-the-shelf (greedy setting) for stance detection.


The best-performing prompt-instruction pair (Appendix~\ref{app:pstance_performance} Tab.~\ref{tab:pstance}) for Donald Trump is 3f-5 ($F_{avg} = 84.43$; greedy), Joe Biden is 4f\_not-7 ($F_{avg} = 84.28$; greeedy), and Bernie Sanders is 3f-8 ($F_{avg} = 82.29$; PMI/AfT).
In terms of the overall $F_{avg}$ score similar to SemEval 2016 Task 6, prompt 3f is the best performer in all decoding strategies. 

Overall, FlanT5-XXL outperforms the SoTA zero-shot baseline---Zero Shot ChatGPT (2023)~\citep{zhang2023investigating} despite never been exposed to an English Twitter stance detection task (Appendix~\ref{app:contamination})---a concern present in ChatGPT~\citep{aiyappa-etal-2023-trust}. 
More impressively, FlanT5-XXL also outperforms the cross-target stance detection baseline (RoBERTa KSCP) which finetunes an LLM on one target and tests on an unseen target---an easier task than a purely zero-shot setting. 
However, it falls short of achieving SoTA performance of LLMs finetuned on the test target (RoBERTa KEPT). The $F_{avg}$ scores per prompt, per instruction, and per target, for P-Stance are shown in Appendix~\ref{app:pstance_greedy},~\ref{app:pstance_pmi},~\ref{app:pstance_aft} for greedy, PMI and AfT decoding respectively.

Unlike on SemEval 2016 Task 6, we find that pre-processing of tweets as suggested by \citet{ghosh2019stance} does not significantly improve the observed performance. 


%
\section{Discussion and Conclusion}
In this work, we have examined the zero-shot stance detection performance of a pre-trained LLM (FlanT5-XXL) against the strongest baselines. 
We show that  FlanT5-XXL shows state-of-the-art performance on SemEval 2016 Task 6B, and comes close to existing state-of-the-art (SoTA) in SemEval 2016 Task 6A and P-stance datasets. 
While choosing baselines, we note the lack of consensus and pervasive oversight on current SoTA, wherein some older models outperform newer models under comparable setups (\S~\ref{sec:baseline}). 
While a recent review~\citep{alturayeif2023systematic} documented most baselines, a rigorous central system tracking the performance of models is missing and is much needed.
Nevertheless, in our literature review, we find that zero-shot stance detection using task-agnostic, off-the-shelf, open-source LLMs on English Twitter datasets used in this work is relatively unexplored. 

We also investigate how and why the model performance is affected by various factors, including the instruction phrasing, the presence of opposition or negation in prompts, decoding strategies, and the presence of test data in the model's training datasets.  
We find that the model is not sensitive to instruction paraphrasing while being negatively affected by the existence of opposition or negation in the prompts.
We consistently find a moderate but significant negative correlation between prompt perplexity and performance which suggests that creating low perplexity prompts may be a useful strategy for zero-shot tasks. 
We show that greedy decoding, the default strategy in most cases, offers competitive performance and that using alternate strategies---PMI or AfT decoding---does not necessarily improve the performance. 
We also notice that, as expected, PMI and AfT decoding strategies show similar performance to each other. 
Future work may explore different $\textbf{b}$ vectors (Eq.~\ref{eq:AfT}) and estimations of $P(y_i)\text{ or } p_{cf}$ (Eq.~\ref{eq:pmi}). 
Most importantly, to ensure that FlanT5-XXL has not been exposed to the SemEval 2016 Task 6 and P-Stance datasets, we analyze its pre-training and instruction tuning data and find no evidence of this. 
In this process, we also identified an over-representation of tokens with a positive connotation, that was used in our prompts, in comparison to their negative counterparts. 
We posit that this may be one of the important reasons why the greedy strategy performs well which aligns with our observations that the model is biased towards outputting tokens with positive connotations.
Lastly, we find that pre-processing tweets improves the LLM's performance on the SemEval 2016 Task 6 dataset but not so much in P-Stance. 

In sum, off-the-shelf FlanT5-XXL shows performance that closely tracks, or even exceeds, SoTA stance detection models. 
Our results demonstrate that the LLM-based zero-shot approach may be all we need for (social media) stance detection task given its strong performance, ease of use, and generalizability. 
Our investigations into the data leakage, prompting, and decoding strategies, may serve as a guide for effectively performing zero-shot stance detection for computational social scientists.

\section{Limitations}

We would like to highlight the following. 
First, while we compare the performance of FlanT5-XXL against the strongest baselines available, we leave it for future work to delve into a similar in-depth performance analysis of the latest open-source LLMs like Llama2 and Alpaca.
We specifically would like to emphasize the need to pay close attention to whether these models have already been exposed to the task's datasets.   
Second, our evaluation is focused on two popular English Twitter datasets---SemEval 2016 Task 6, and P-Stance. 
It would be interesting and important future work to test the performance of LLMs (specifically FlanT5-XXL) when there is a distributional shift of test data, on multi-lingual and multi-target stance detection, and in datasets where the target is not specified. 
Third, although we observe the models' sensitivity to negations and oppositions, the presence of positivity bias in both model output and training data, and significant negative correlations between prompt perplexity and model performance, none of these provide causal explanations for the observed performance and sensitivity.
Lastly, LLMs are constrained by their context length and the need for GPU resources. 
The latter also has a significant environmental impact.
For instance, FlanT5-XXL requires 40GB of GPU RAM and has a limited context length of 512 tokens. 
This may limit the length of input text from which the stance needs to be inferred. 
However, with modern hardware and optimization, coupled with modern LLMs' capability of handling a much larger context this limitation may disappear soon.
\section*{Ethics Statement}
In this work, we presented the results obtained by a large language model (LLM)---FlanT5-XXL. While the bias of FlanT5-XXL to output tokens with positive connotation over negative has been explored in this work, other biases like gender, and race, have not been explored but may be in play when the LLM assigns stances to tweets.
Given the trend of newer LLMs being closed-source, documentation of the data of training data is sparse (if any) and it is becoming harder to properly quantify such biases and verify the existence of test data in the training set.
This calls for more efforts in designing experiments to quantify the impact of training data and its features, on various NLP tasks.

Lastly, using LLMs requires powerful GPUs which consume significant energy and contribute to climate change, whose effects are more starkly felt by marginalized groups and countries. 
We note that zero-shot setting consumes significantly fewer resources compared to pre-training or fine-tuning.
We also note that access to powerful GPUs is a barrier to research and/or to reproduce the results in this paper.
\section*{Acknowledgements}
%
We acknowledge the computing resources at Indiana University and thank NVIDIA for their GPU resources. One of the authors was supported by DARPA under contract HR001121C0168.

\bibliography{anthology,custom}
\bibliographystyle{acl_natbib}
\clearpage
\appendix
\clearpage
\onecolumn
\section{SemEval 2016 Task 6A Dataset}~\label{app:task6}
\begin{table*}[htbp]
\centering
\begin{tabular}{|c|c|c c c|c|c c c|c|}
\hline
\multicolumn{1}{|c|}{\multirow{2}{*}{Target}} & \multicolumn{4}{c|}{Training Set (\%)} & \multicolumn{4}{|c|}{Test Set (\%)} & \multicolumn{1}{c|}{\multirow{2}{*}{Total}} \\ 
\cline{2-9}
      & Train & Favor & Against & None & Test & Favor & Against & None &  \\
\hline
AT     & 513 & 17.9  & 59.3  &  22.8 & 220 &  14.5  & 72.7    & 12.7 & 733 \\
CC   & 395 & 53.7  & 3.8     & 42.5 & 169 &  72.8  & 6.5     & 20.7 & 564 \\
HC    & 689 & 17.1  & 57.0    & 25.8 & 295 & 20.4  & 64.2    & 15.4 & 984 \\
FM    & 664 & 31.6  & 49.4    & 19.0 & 285 & 15.3  & 58.3    & 26.4 & 949 \\
LA    & 653 & 18.5  & 54.4    & 27.1 & 280 & 16.4  & 67.5    & 16.1 & 933 \\
\hline
Total & 2914 & 25.8  & 47.9    & 26.3 & 1249 & 24.3  & 57.3    & 18.4 & 4163 \\
\hline
\end{tabular}
\caption{Percentage of tweets in the training and test sets for each target in Task 6A of SemEval 2016}
\label{table:Task6A_data}
\end{table*}

\begin{table*}[htbp]
\centering
\begin{tabular}{|c|c|c c c|c|c c c|c|}
\hline
\multicolumn{1}{|c|}{\multirow{2}{*}{Target}} & \multicolumn{4}{c|}{Training Set (\%)} & \multicolumn{4}{c|}{Test Set (\%)} & \multicolumn{1}{c|}{\multirow{2}{*}{Total}} \\ 
\cline{2-9}
      & Train & Favor & Against & None & Test & Favor & Against & None &  \\
\hline
DT     & - & -  & -  &  - & 707 &  21  & 42.3   & 36.7 & 707 \\
\hline
\end{tabular}
\caption{Percentage of tweets in the training and test sets for each target in Task 6B of SemEval 2016}
\label{table:Task6B_data}
\end{table*}
\begin{table*}[htbp]
\centering
\begin{tabular}{c*{6}{c}}
\toprule
Target & AT & CC & DT & FM & HC & LA \\
\midrule
\# Tweets & 935,181 & 208,880 & 78,156 & 144,166 & 238,193 & 113,193 \\
\bottomrule
\end{tabular}
\caption{Number of tweets by target in the domains corpurs of SemEval 2016 Task 6.}
\label{tab:domain_corpus}
\end{table*}

%
\clearpage
\onecolumn
\newpage

\section{P-Stance Dataset}~\label{app:pstance}
\begin{table*}[h!]
\centering
\begin{tabular}{|c|c c|c c|c c|c|}
\hline
\multicolumn{1}{|c|}{\multirow{2}{*}{Target}} & \multicolumn{2}{c|}{Training Set} & \multicolumn{2}{c|}{Validation Set} & \multicolumn{2}{c|}{Test Set} & \multicolumn{1}{c|}{\multirow{2}{*}{Total}} \\ 
\cline{2-7}
      & Favor & Against & Favor & Against & Favor & Against &  \\
\hline
Trump     & 2,937 & 3,425  & 365  &  430 & 361 &  435  & 7,953 \\
Biden   & 2,552 & 3,254  & 328 & 417 & 337 &  408  & 7,296 \\
Sanders   & 2,858 & 2,198  & 350  & 284 & 343 & 292  & 6,325 \\
\hline
\end{tabular}
\caption{Number of tweets in the training, validation, and test sets for each target in P-Stance}
\label{tab:pstance_data}
\end{table*}
\clearpage
\twocolumn
\section{Learning from  experiments with other LLMs}~\label{app:preliminary}
A major advantage of FlanT5-XXL over bigger, but non-instruction-tuned models like OPT-175, and BLOOM-176B is its ability to follow instructions rigorously. 
This was also verified in our preliminary experiments with OPT-175B and BLOOM-176B on the SemEval dataset where it was hard to make these models produce desired output labels and not produce garbage text from the perspective of stance detection. 
Across a range of settings, we observe that FlanT5-XXL always outputs one of the options in the instructions (Tab.~\ref{tab:prompts}) and no further parsing or post-processing of the response was necessary, unlike OPT-175, and BLOOM-176 which often gave verbose and irrelevant outputs. 
Such models, as a post-processing step, thus required us to compute probabilities of the labels (`Favor,' `Against,' and `Neutral') given the context and choose the one among them, with, say, the higher probability (greedy decoding), making it more technical to use than FlanT5-XXL---an undesirable property from the lens of off-the-shelf large scale usage. 
\section{Instructions}~\label{app:instrutions}
9 paraphrases of ``Answer the question with \{opt-1\}, \{opt-2\}, or \{opt-3\}." obtained by ChatGPT, where \{opt-1\}, \{opt-2\}, and \{opt-3\} are indicated in the `Instruction Outputs' column in Tab.~\ref{tab:prompts}.

\begin{enumerate}
    \item ``Your response to the question should be either \{opt-1\}, \{opt-2\}, or \{opt-3\}."
    \item ``The question needs to be answered with either \{opt-1\}, \{opt-2\}, or \{opt-3\}."
    \item ``Kindly provide your answer to the question in the format of \{opt-1\}, \{opt-2\}, or \{opt-3\}."
    \item ``To provide your response to the question, you need to select \{opt-1\}, \{opt-2\}, or \{opt-3\}."
    \item ``The question requires a response in the form of \{opt-1\}, \{opt-2\}, or \{opt-3\}."
    \item ``Kindly respond to the question using \{opt-1\}, \{opt-2\}, or \{opt-3\}."
    \item ``Please select one of \{opt-1\}, \{opt-2\}, or \{opt-3\} when answering the question."
    \item ``In order to answer the question, you must select either \{opt-1\}, \{opt-2\}, or \{opt-3\}."
    \item ``To provide your response to the question, please select \{opt-1\}, \{opt-2\}, or \{opt-3\}." 
\end{enumerate}
\section{Evaluation Metrics}~\label{app:evaluation_metrics}

\begin{equation}
    Precision = \frac{TP}{TP + FP}
\end{equation}
\begin{equation}
    Recall = \frac{TP}{TP + FN}
\end{equation}
\begin{equation}
    F = \frac{2\cdot Precision\cdot Recall}{Precision + Recall}
\end{equation}
\begin{equation}
    F_{avg} = \frac{F_\text{favor} + F_\text{against}}{2}
\end{equation}
\section{Approaches used by baselines}~\label{app:baselines}
\paragraph{SemEval 2016 Task 6A}
(1) LibSVM (2017): This is a supervised model trained only on the task's training data. This model uses 36,339 lexical features divided into 18 feature sets comprising of sentiment, emojis, n-grams, etc.~\citep{hacohen2017stance}, and despite its age, still performs better than more recent models, 
(2) PE-HCN (2020)~\citep{zhao2020pretrained}: This approach, currently the state of the art to the best of our knowledge, fine-tuned RoBERTA on the task's training data, and the resulting embeddings are then used to train a hierarchical capsule network architecture functioning on a dynamic routing algorithm~\citep{sabour2017dynamic}, (3) TimeLM (2022): This is a BERT model which is periodically fine-tuned on latest tweets collected in a quarterly time intervals, and further fine-tuned on the task's training data. This work shows that performance degradation is higher when using a model fine-tuned on past tweets that is evaluated on more recent tweets than a model that is fine-tuned on present tweets performing evaluated on older tweets. We thus use the stance detection results by the latest TimeLM model (2022) since the original work showed this to be the best in stance detection compared to older TimeLM models~\citep{loureiro2022timelms}, and 
(4) ChatGPT (2023): A recent decoder-only large language based on GPT3 and fine-tuned using queries submitted to the OpenAI platform,  which raises questions about contaminated fine-tuning data~\citep{aiyappa-etal-2023-trust} and its evaluation. However, recent work has evaluated its performance on the stance detection task in a zero-shot setting~\citep{kocon2023chatgpt}.

\paragraph{SemEval 2016 Task 6B}
(1) BiLSTM (2016): A weakly-supervised approach that first automatically labels the domain corpus (Appendix~\ref{app:task6} Tab.~\ref{tab:domain_corpus}) of task 6B using heuristic rules like the presence of regular expressions, which is then used to train a BiLSTM model~\citep{augenstein2016stance}, 
(2) SVM (2017): A two-stage classification system that first uses subjectivity features of language (like the presence/absence of adjectives) to classify tweets as neutral/non-neutral, and next, the non-neutral tweets are classified as positive vs. negative using features lexical features like sentiment, n-grams, etc.~\citep{dey2017twitter}, 
(3) POLITICS (2022): Fine-tuned RoBERTa on additional objectives on a newly curated dataset, and further on Task 6A's data to build a language model which can extract ideological and stance differences~\citep{liu2022politics}, and 
(4) PT-HCL (2022): A fine-tuned BERT model on task 6A data that extracts target-specific and target-invariant features and uses noise-contrast learning to improve zero-shot performance~\citep{liang2022zero}. 

\paragraph{P-Stance}
(1) RoBERTA KSCP (2022): This method fine-tuned a RoBERTa model in a way that the target is made to be a part of the prompt template, along with the original statement, during fine-tuning. The fine-tuning process then involves predicting masked tokens in the input prompt and the output (stance) labels. 
Including the target as a part of the prompt is argued to ``stimulate'' internal knowledge acquired by the LLM during training and assist it in assigning stances in cases where the target is not explicitly mentioned in the original statement ~\citep{zheng2022knowledge}. 
This is called RoBERTa Knowledge Stimulated Contrastive Prompting (RoBERTA KSCP). However, the work reports a drop in performance in a cross-target setting---when the fine-tuing is done on one target (say, `Trump'; called `RoBERTa KSCP trump') and tested on a different target (say, `Biden'). 
This cross-target setup is closer to, but not the same as our zero-shot setup. 
(2) RoBERTa KEPT (2023): This method fine-tunes RoBERTa on the task's training data and an additional data set created by substituting the targets and stance labels in the training data with their conceptual/verbal synonyms, thereby enhancing the knowledge of the LLM---RoBERTa Knowledge Enhanced Prompt Tuning (KEPT)~\citep{huang2023knowledge}.
Specifically, the pipeline uses a (i) verbaliser (a semantic graph) to build a better mapping of the predicted word by the LLM to ground-truth labels, (ii) concept graph to supplement the target with a synonymous concept during fine-tuning, and (iii) a topic model to learn the representation of hashtags in the tweets to help the LLM understand hashtags. 
This work does not present scores for the target ``Bernie Sanders.'' 
(3) COT ChatGPT (2023) This work evaluates ChatGPT's performance in a Chain-of-Thought setting~\citep{zhang2023investigating} 
(4) COT ChatGPT (2023) This tests the zero-shot capability of ChatGPT~\citep{zhang2023investigating}---the only baseline that offers a fair comparison to our work. 
This is also the current SoTA in a zero-shot setting.
\section{Commentary on the baselines claimed to be state of the art in previous works}~\label{app:questioning_baselines}
We wish to highlight that ``state-of-the-art'' baselines used in studies often do not contain the actual state-of-the-art. 
For instance, on Task 6A, \citet{barbieri2020tweeteval} shows fine-tuned RoBERTa, outperforms ($F_{avg} = 71.0$) chosen baselines like SVM, LSTM, and FastText. Similarly,~\citet{loureiro2022timelms} fine-tuned BERT to obtain $F_{avg} = 72.6$, and \citet{liu2022politics} show $F_{avg} = 70.09.$
However, before any of these works, systems by \citet{hacohen2017stance}, and \citet{zhao2020pretrained} had already achieved $F_{avg}$ scores of $77.11$ and $78.43$, respectively, which were never considered as baselines to compare against. 

In Task 6B, \citet{liang2022zero} show an $F_{avg}=50.1$ but do not compare against baselines like \citet{augenstein2016stance} or the model by \citet{dey2017twitter} with $F_{avg}$ scores of $58.03$ and $61.57$ respectively.

\section{Perplexity Analysis}~\label{app:perp}
Based on the hypothesis that a language model answers better, a question it understands better~\citep{gonen2022demystifying}, we check the correlation between the average (across all targets) per prompt ID, per instruction perplexity of the prompts used on the test data of SemEval 2016 Tasks 6A and 6B, and the $F_{avg}$ score across all the targets of the respective tasks. We also check the same correlation using context-free prompts (no $\langle tweet \rangle$ included in Tab.~\ref{tab:prompts}). We repeat the same on individual targets of the P-Stance dataset. 

The unconditional perplexity of the prompt in an encoder-decoder model is loosely defined in contrast to a decoder model. To calculate the perplexity, we feed an empty string, ` ' into the encoder of FlanT5-XXL and calculate the cross-entropy of the decoder generating the desired prompt. Perplexity\footnote{https://leimao.github.io/blog/Entropy-Perplexity/} is defined as
\begin{equation}
    PP({\tilde{p}},q) = b^{H({\tilde{p}}, q)} = b^{\mathbb{E}_{\tilde{p}}[log_b q]}
\end{equation}\label{eq:perplexity }
where $H({\tilde{p}}, q) = - \sum_{i=1}^{n} \tilde{p}(x_i) \log_b q(x_i)$ 
is the cross entropy, $\tilde{p}$ is the one hot encoding of the token $x_i$ in a sentence of length $n$, and $q$ is the predicted probability of the token by FlanT5-XXL.
We use the lmppl python package\footnote{https://github.com/asahi417/lmppl} to calculate the perplexities of the texts.

While this hypothesis has its limitations, for instance, a totally unrelated task might have a prompt with perplexity much less than any of the prompts for the stance detection task, we argue that it may nevertheless provide some explanation for the observed performance of FlanT5-XXL. 

\begin{figure*}[t]
    \centering
    \includegraphics[width=1\textwidth, trim={0 0in 0 0in}]{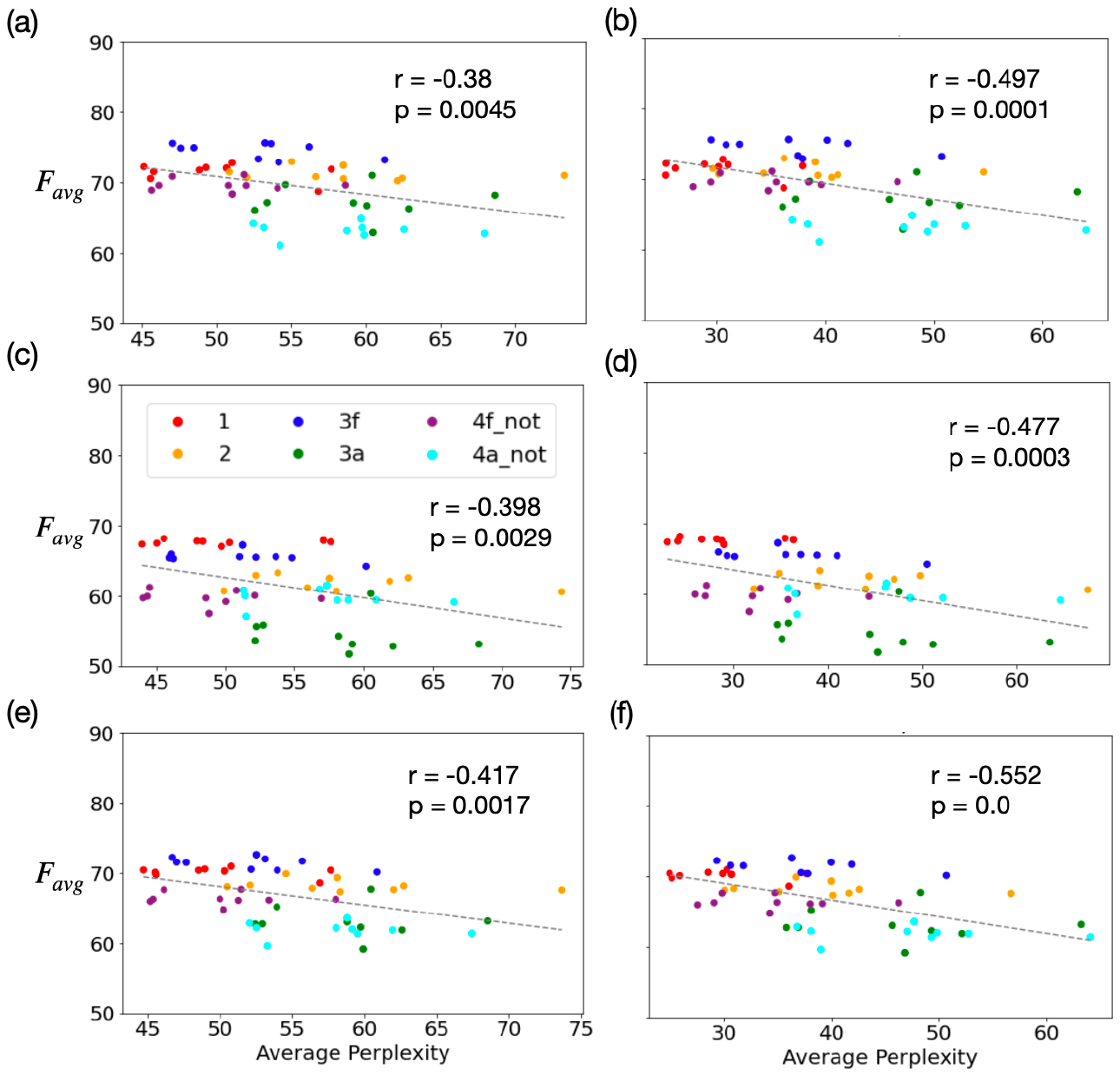}
    \caption{Correlation between prompt perplexity, per prompt ID, per instruction, and $F_{avg}$ scores (from greedy) across targets of SemEval 2016 Task 6. (a,c,e) Prompts with the $\langle tweet \rangle$ object. (b,d,f) Prompts without the $\langle tweet \rangle$  object---context-free prompt.  (a,b) Task A, (c,d) Task B, (e,f) Task A+B. Correlation coefficients are indicated by $r$ and p-values by $p$.}\label{fig:perplexities}
\end{figure*}

In Fig.~\ref{fig:perplexities}, we see a moderate but statistically significant correlation between prompt perplexities and $F_{avg}$ scores on the SemEval 2016 Task 6. Across the tasks and prompt types (context-free, or with context), we see that Prompt ID 1 and 3f fall to the top left of the space which means that FlanT5-XXL \textit{understands} these prompts better and performs well on the task with these prompts. This may explain the superior performance of prompts 1 and 3f relative to the other prompts. Similar to the argument made in \S\ref{sec:results}, we observe that prompts with opposition (3a), and negation (4a\_not) fall in the bottom right of the plot, with high perplexity and poor performance. We also note the difference in perplexities of the 3f prompt and its negation (4f\_not) with the latter having higher perplexity and poorer performance.   

Using an independent sample t-test, we also observe a significant difference (p<0.005) between the average perplexities of context-free prompts across Tasks 6A and 6B and the actual prompts (Fig.~\ref{fig:perplexities}, e,f), with the latter having higher perplexities, indicating that it is harder for the language model to understand tweets.

\begin{figure*}[t]
    \centering    \includegraphics[width=1\textwidth, trim={0 1in 0 1in}]{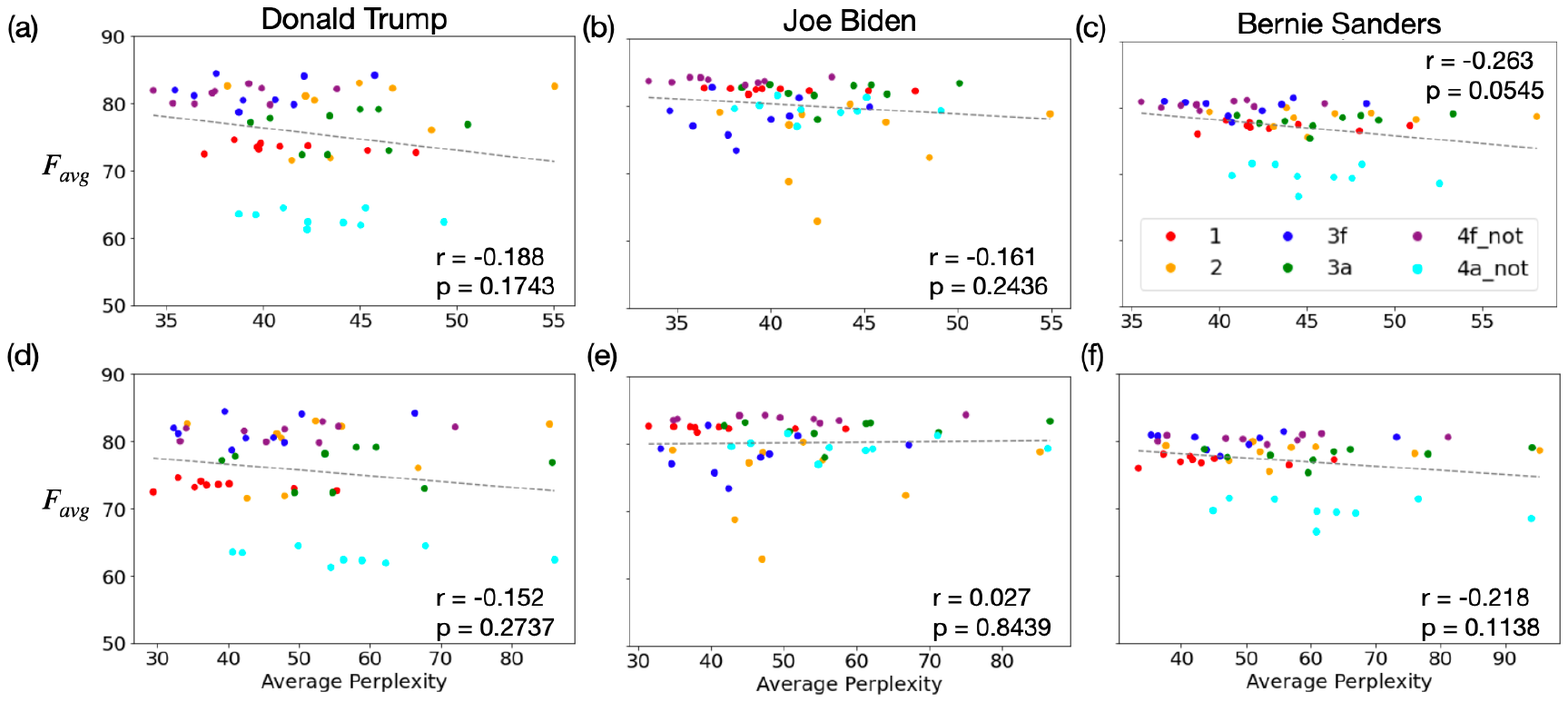}
    \caption{Correlation between prompt perplexity, per prompt ID, per instruction, and $F_{avg}$ scores (from greedy) for each target in the P-Stance dataset. (a,d) Donald Trmup, (b,e) Joe Biden, (c,f) Bernie Sanders. (a,b,c) Prompts with the $\langle tweet \rangle$ object. (d,e,f) Prompts without the $\langle tweet \rangle$  object---context-free prompt. Correlation coefficients are indicated by $r$ and p-values by $p$. }\label{fig:perplexities_pstance}
\end{figure*}

In Fig.~\ref{fig:perplexities_pstance} we see no significant correlation between prompt perplexities and $F_{avg}$ scores on individual targets in the P-Stance dataset. This agrees with our finding that the difference in performance between prompts is lesser in the P-Stance dataset (see Fig.~\ref{fig:pstance}) compared to the SemEval 2016 Task 6 dataset (see Fig.~\ref{fig:taskAB}). Nevertheless, we do observe the two best-performing prompts (3f and 4f\_not) occupy the top left of the plots---low perplexity, high $F_{avg}$. A surprising finding we are yet to understand is the perplexities of context-free prompts being higher than that with context.  

In sum, we find that measuring the perplexity of prompts might serve as a guiding principle to generate better prompts and in turn improve the model's performance on the task.
\clearpage
\onecolumn
\section{Sensitivity to Instructions}~\label{app:instruction_sensitivity}
\begin{table*}[ht]
\begin{center}
\setlength\tabcolsep{1.5pt} 
\begin{tabular}[t]{ccccccc}
\hline
PromptID $\backslash$Target & AT & CC & LA & FM & HC & $F_{avg}$\\
\hline
\multirow{1}{4em}{\centering 1} & 
\multirow{1}{4em}{\centering 0.51}
& \multirow{1}{4em}{\centering 1.30}& \multirow{1}{4em}{\centering 0.31} & \multirow{1}{4em}{\centering 0.14} & \multirow{1}{4em}{\centering 0.34} &  \multirow{1}{4em}{\centering 0.40} 
\\
\multirow{1}{4em}{\centering 2} & 
\multirow{1}{4em}{\centering 0.46}
& \multirow{1}{4em}{\centering 0.83}& \multirow{1}{4em}{\centering 0.44} & \multirow{1}{4em}{\centering 0.64} & \multirow{1}{4em}{\centering 0.49} &  \multirow{1}{4em}{\centering 0.31} 
\\
\multirow{1}{4em}{\centering \textbf{3f}} & 
\multirow{1}{4em}{\centering \textbf{0.43}}
& \multirow{1}{4em}{\centering \textbf{1.09}}& \multirow{1}{4em}{\centering \textbf{0.65}} & \multirow{1}{4em}{\centering \textbf{0.76}} & \multirow{1}{4em}{\centering \textbf{0.92}} &  \multirow{1}{4em}{\centering \textbf{0.37}} 
\\
\multirow{1}{4em}{\centering 3a} & 
\multirow{1}{4em}{\centering 1.41}
& \multirow{1}{4em}{\centering 1.05}& \multirow{1}{4em}{\centering 0.92} & \multirow{1}{4em}{\centering 0.68} & \multirow{1}{4em}{\centering 1.20} &  \multirow{1}{4em}{\centering 0.77} 
\\
\multirow{1}{4em}{\centering 4f\_not} & 
\multirow{1}{4em}{\centering 0.37}
& \multirow{1}{4em}{\centering 0.62}& \multirow{1}{4em}{\centering 0.46} & \multirow{1}{4em}{\centering 0.46} & \multirow{1}{4em}{\centering 0.77} &  \multirow{1}{4em}{\centering 0.29} 
\\
\multirow{1}{4em}{\centering 4a\_not} & 
\multirow{1}{4em}{\centering 0.65}
& \multirow{1}{4em}{\centering 0.64}& \multirow{1}{4em}{\centering 0.49} & \multirow{1}{4em}{\centering 0.36} & \multirow{1}{4em}{\centering 0.42} &  \multirow{1}{4em}{\centering 0.36} \\
\hline
\end{tabular}
\caption{Standard mean errors of $F_{avg}$ scores from greedy for each prompt ID across the 9 instructions, on SemEval 2016 Task 6A using greedy decoding. The best performer in the task, from Tab.~\ref{tab:taskA} is indicated using boldface.}~\label{tab:sme_taskA}
\end{center}
\end{table*} 
\begin{table*}[ht]
\begin{center}
\setlength\tabcolsep{1.5pt} 
\begin{tabular}[t]{cc}
\hline
Prompt ID $\backslash$Target & $F_{avg}$\\
\hline
\multirow{1}{8em}{\centering \textbf{1}} & 
\multirow{1}{4em}{\centering \textbf{0.11}}
\\
\multirow{1}{8em}{\centering 2} & 
\multirow{1}{4em}{\centering 0.35}
\\
\multirow{1}{8em}{\centering \textit{3f}} & 
\multirow{1}{4em}{\centering \textit{0.26}}
\\
\multirow{1}{8em}{\centering 3a} & 
\multirow{1}{4em}{\centering 0.85}
\\
\multirow{1}{8em}{\centering 4f\_not} & 
\multirow{1}{4em}{\centering 0.35}
\\
\multirow{1}{8em}{\centering 4a\_not} & 
\multirow{1}{4em}{\centering 0.44}\\
\hline
\end{tabular}
\caption{Standard mean errors of $F_{avg}$ scores from greedy for each prompt ID across the 9 instructions, on SemEval 2016 Task 6B using greedy decoding. The best performer in the task, from Tab.~\ref{tab:taskB} is indicated using boldface and the second best performer, which happens to be the best performer in Task 6A is shown in italics. }~\label{tab:sme_taskB}
\end{center}
\end{table*} 
\begin{table*}[ht]
\begin{center}
\setlength\tabcolsep{1.5pt} 
\begin{tabular}[t]{ccccccc }
\hline
Prompt ID$\backslash$Dec. & Greedy (A) & PMI (A) & AfT (A) & Greedy (B) & PMI (B) & AfT (B)\\
\hline
\multirow{1}{8em}{\centering 1} & 
\multirow{1}{4em}{\centering 0.40} & 
\multirow{1}{4em}{\centering 1.10} & 
\multirow{1}{4em}{\centering 1.10} & 
\multirow{1}{4em}{\centering 0.11} & 
\multirow{1}{4em}{\centering 0.72} & 
\multirow{1}{4em}{\centering 0.65}
\\
\multirow{1}{8em}{\centering 2} & 
\multirow{1}{4em}{\centering 0.31} & 
\multirow{1}{4em}{\centering 0.53} & 
\multirow{1}{4em}{\centering 0.53} & 
\multirow{1}{4em}{\centering 0.35} & 
\multirow{1}{4em}{\centering 0.77} & 
\multirow{1}{4em}{\centering 0.69}
\\
\multirow{1}{8em}{\centering 3f} & 
\multirow{1}{4em}{\centering 0.37} & 
\multirow{1}{4em}{\centering 1.47} & 
\multirow{1}{4em}{\centering 1.47} &
\multirow{1}{4em}{\centering 0.26} & 
\multirow{1}{4em}{\centering 0.76} & 
\multirow{1}{4em}{\centering 1.75} 
\\
\multirow{1}{8em}{\centering 3a} & 
\multirow{1}{4em}{\centering 0.77} & 
\multirow{1}{4em}{\centering 1.17} & 
\multirow{1}{4em}{\centering 1.17} &
\multirow{1}{4em}{\centering 0.85} & 
\multirow{1}{4em}{\centering 0.69} & 
\multirow{1}{4em}{\centering 0.62} 
\\
\multirow{1}{8em}{\centering 4f\_not} & 
\multirow{1}{4em}{\centering 0.29} & 
\multirow{1}{4em}{\centering 0.26} & 
\multirow{1}{4em}{\centering 0.26} &
\multirow{1}{4em}{\centering 0.35} & 
\multirow{1}{4em}{\centering 0.35} & 
\multirow{1}{4em}{\centering 0.32} 
\\
\multirow{1}{8em}{\centering 4a\_not} & 
\multirow{1}{4em}{\centering 0.36} & 
\multirow{1}{4em}{\centering 0.58} & 
\multirow{1}{4em}{\centering 0.58} &
\multirow{1}{4em}{\centering 0.44} & 
\multirow{1}{4em}{\centering 0.65} & 
\multirow{1}{4em}{\centering 0.86}
\\
\hline
\end{tabular}
\caption{Standard mean errors of $F_{avg}$ scores for each prompt ID across the 9 instructions, on Task 6A and 6B, comparing greedy decoding to PMI decoding.}~\label{tab:sme_greed_pmi}
\end{center}
\end{table*} 
\begin{table*}[ht]
\begin{center}
\setlength\tabcolsep{1.5pt} 
\begin{tabular}[h!]{cccc}
\hline
P. ID $\backslash$Target & DT & JB & BS \\
\hline
\multirow{1}{8em}{\centering 1} & 
\multirow{1}{8em}{\centering 0.22 (0.31) [0.31]} & 
\multirow{1}{8em}{\centering 0.1 (0.26) [0.26]} & 
\multirow{1}{8em}{\centering 0.21 (0.28) [0.28]} 
\\
\multirow{1}{8em}{\centering 2} & 
\multirow{1}{8em}{\centering 1.55 (0.83) [0.83]} & 
\multirow{1}{8em}{\centering 1.96 (0.68) [0.68]} & 
\multirow{1}{8em}{\centering 0.45 (0.76) [0.76]} 
\\
\multirow{1}{8em}{\centering 3f} & 
\multirow{1}{8em}{\centering 0.7 (0.32) [3.91]} & 
\multirow{1}{8em}{\centering 0.94 (0.31) [4.63]} & 
\multirow{1}{8em}{\centering 0.4 (0.22) [0.22]} 
\\
\multirow{1}{8em}{\centering 3a} & 
\multirow{1}{8em}{\centering 0.94 (0.5) [0.5]} & 
\multirow{1}{8em}{\centering 0.56 (0.2) [0.2]} & 
\multirow{1}{8em}{\centering 0.38 (0.48) [0.49]} 
\\
\multirow{1}{8em}{\centering 4f\_not} & 
\multirow{1}{8em}{\centering 0.39 (0.16) [0.16]} & 
\multirow{1}{8em}{\centering 0.14 (0.11) [0.11]} & 
\multirow{1}{8em}{\centering 0.17 (0.13) [0.13]} 
\\
\multirow{1}{8em}{\centering 4a\_not} & 
\multirow{1}{8em}{\centering 0.38 (0.8) [0.86]} & 
\multirow{1}{8em}{\centering 0.45 (0.22) [0.22]} & 
\multirow{1}{8em}{\centering 0.54 (0.33) [0.33]}\\
\hline
\end{tabular}
\caption{Standard mean errors of $F_{avg}$ scores for each prompt ID across the 9 instructions, on P-Stance dataset using greedy decoding. The standard mean errors from PMI and AfT decoding are shown in round and square brackets respectively. Prompt 3f and 4f\_not were the best performers in the task as seen in Tab.~\ref{tab:pstance}.
}~\label{tab:sme_pstance}
\end{center}
\end{table*} 
\newpage
\clearpage
\twocolumn
\section{Sensitivity to Decoding strategy}~\label{app:PMI}
A potential reason for the observed performance degradation of PMI and AfT decoding in contrast to greedy decoding (when the opposite was expected~\citep{holtzman2021surface}) in SemEval 2016 Task 6 dataset could be that the model is biased towards the answer which is the majority class in the test data (i.e `against'; See Tab.~\ref{table:Task6A_data},~\ref{table:Task6B_data}). 
Hence PMI and AfT decoding, which helps to reduce such biases---stemming from pre-training and instruction/fine-tuning---in the LLM outputs, especially in classification tasks, perhaps reflects a truer evaluation of the model. 
For example, in an extreme case, if all of the test data had the stance ``against'' and the model is biased towards outputting this label, then compared to an unbiased model, a biased model would show better performance if it were to use greedy decoding. 
In contrast, the PMI or AfT decoding, which claims to reduce such biases~\citep{holtzman2021surface,zhao2021calibrate}, might perform poorer than greedy.
The fact that such a stark degradation in performance doesn't appear on the P-stance dataset, which is more balanced, also supports our hypothesis.

To test this hypothesis, we gauged FlanT5-XXL's bias by first posing the same prompts shown in Tab.~\ref{tab:prompts} without actually feeding in the $\langle tweet \rangle$.

\begin{figure*}[t]
    \centering
    \includegraphics[width=1\textwidth, trim={0 2in 0 2.5in}]{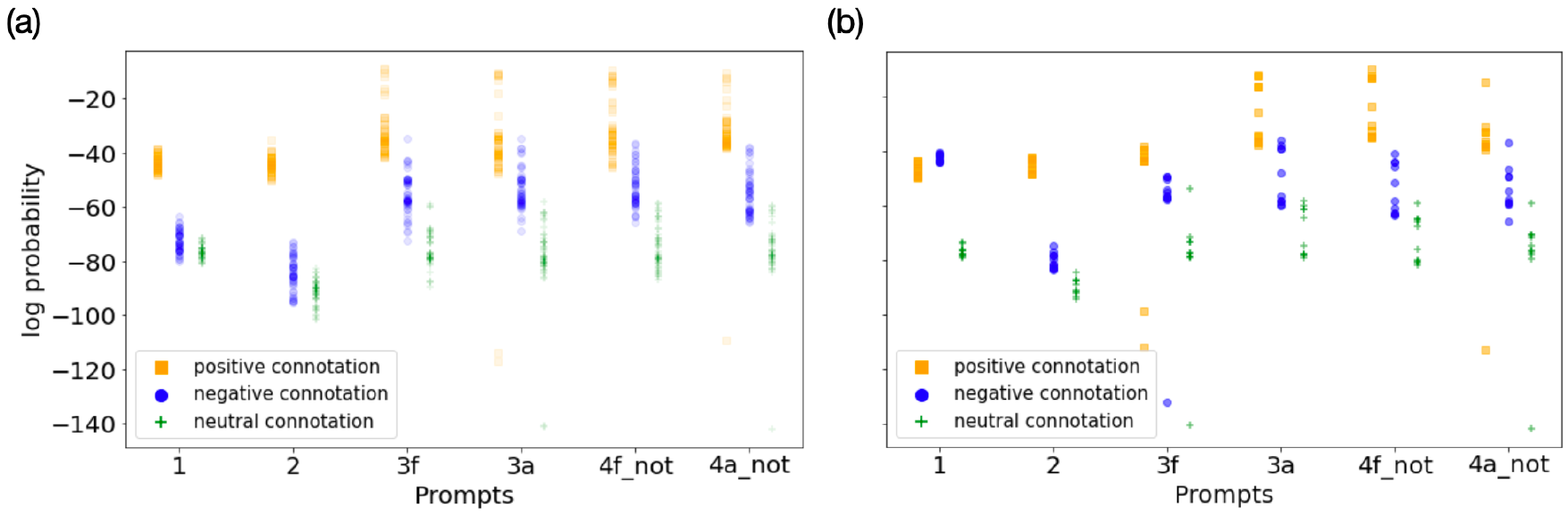}
    \caption{Probability of FlanT5-XXL outputting a label with a positive, negative, or neutral connotation in the SemEval 2016 Task 6, in a context-free setting---a setting where the $\langle tweet \rangle$ item in Tab.~\ref{tab:prompts} is not fed into the model during inference. We see that the model is biased towards labels with positive connotations regardless of prompts and instruction in both (a) Task 6A, and (b) Task 6B. Each point represents a (prompt, instruction, target) tuple (see Tab.~\ref{tab:prompts}).}\label{fig:cf_semeval}
\end{figure*}
\begin{figure*}[t]
    \centering
    \includegraphics[width=1\textwidth, trim={0 3in 0 3in}]{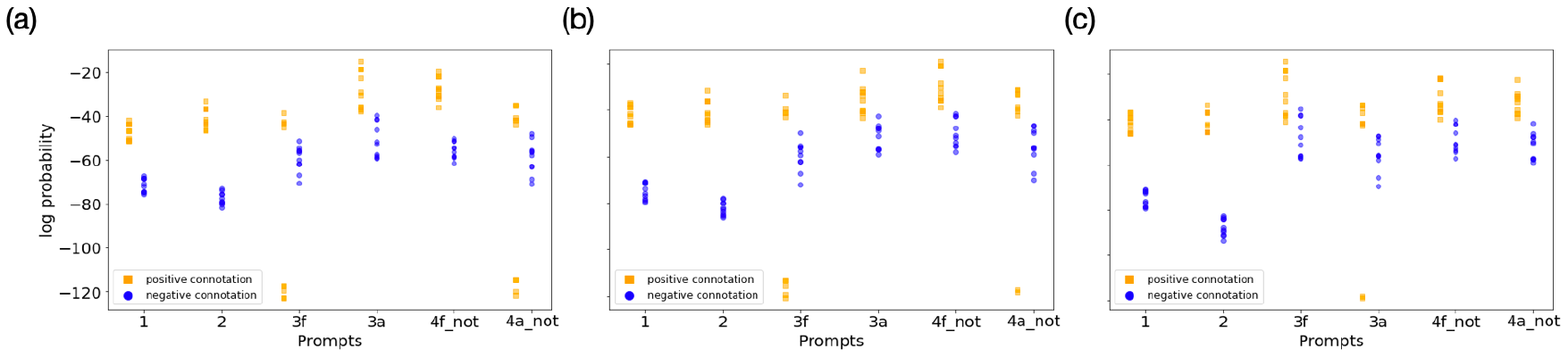}
    \caption{Probability of FlanT5-XXL outputting a label with a positive, or negative in the P-Stance dataset, in a context-free setting---a setting where the $\langle tweet \rangle$ item in Tab.~\ref{tab:prompts} is not fed into the model during inference. We see that the model is on average biased towards labels with positive connotations regardless of prompts and instruction. Each point represents a (prompt, instruction) pair (see Tab.~\ref{tab:prompts}) (a) Donald Trump (b) Joe Biden, and (c) Bernie Sanders.}\label{fig:cf_pstance}
\end{figure*}

Figs.~\ref{fig:cf_semeval} and~\ref{fig:cf_pstance} show that on the SemEval 2016 Task 6, and P-Stance datasets respectively, Flant5-XXL is biased to labels with positive connotations across different prompt types. 
That is for prompt 1 (see.~\ref{tab:prompts}), FlanT5-XXL is biased towards outputting ``positive,'' for prompt 2, ``favor'', and for the rest of the prompts, ``true.'' 
The ``against'' label is the majority label in the test sets of SemEval 2016 Task 6 (see Tabs.~\ref{table:Task6A_data},~\ref{table:Task6B_data}), and P-Stance (see Tab.~\ref{tab:pstance_data})---though to a lesser extent. 
Therefore, from our hypothesis, we would expect that for prompts 3a and 4f\_not, the bias of the model would result in greedy performing better than PMI or AfT. This is always true in Fig.~\ref{fig:taskAB} and mostly true in Fig.~\ref{fig:pstance}. 
However, this still does not explain why PMI and AfT perform worse than greedy in the other prompts when our hypothesis expects the opposite. 
For instance, in prompt 3f (``The statement is in favor of $\langle tweet \rangle$''), the model is biased towards outputting ``true.'' 
In the greedy setting, the model is then expected to assign the label ``favor''---the minority label in the test datasets---to a majority of test instances while PMI and AfT should correct this bias and hence result in a better performance than greedy. We leave further exploration of this for future work. 
Particularly, exploring whether the ordering of the instruction options in~\ref{app:instrutions} can explain this is worth exploring~\cite{pezeshkpour2023large}. 

To gauge the source of this bias of FlanT5-XXL to output positive tokens, we probe into the pre-training and instruction tuning data of the model (see Appendix~\ref{app:positivitiy_bias}).
\clearpage
\newpage
\twocolumn
\section{Source of positivity bias}~\label{app:positivitiy_bias}
As a first step towards understanding the observed (different) performances of different prompts, we took a ground-up approach and extracted the unigram counts of certain key tokens (which distinguish the prompts) and the number of unique bigrams they are part of in the training (Appendix.~\ref{app:C4}), and instruction-tuning data of FlanT5 available on hugging face.\footnote{https://huggingface.co/datasets/allenai/c4}$^,$\footnote{https://huggingface.co/datasets/SirNeural/flan\_v2} We do this by extracting text, and finding contiguous alphanumeric characters, including uppercase and lowercase letters, digits, and underscores, using the regular expression $`\backslash w+'$.


\begin{figure*}[t]
    \centering
    \includegraphics[width=1\textwidth, trim={0 2in 0 2in}]{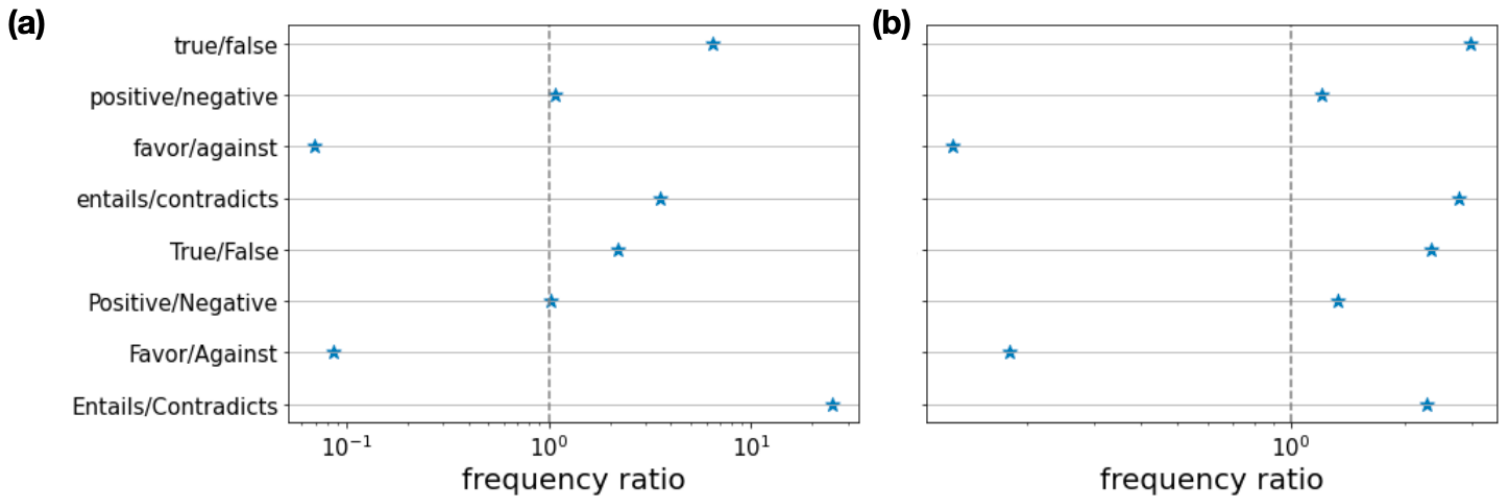}
    \caption{The (a) unigram (b) unique bigrams ratio of the frequency of a token with positive connotation and its antonym in the training data + instruction-tuning data. The y-axis indicates the token pair. We see that except for the favor-against pair, the number of unigrams or unique bigrams a token with a positive connotation participates in is greater than the its negative counterpart. }\label{fig:uni_bi}
\end{figure*}

Fig.~\ref{fig:uni_bi} shows a ``positivity bias''  in the datasets~\citep{dodds2015human}---tokens with positive connotations appear more often (based on counting unigrams) than tokens with negative connotations. 
This may explain why the models are biased towards outputting positive labels (Appendix~\ref{app:PMI}). However, the token ``against'' (``Against'') appears more often than ``favor'' (``Favor'')  both in terms of unigram and unique bigrams. Perhaps this may be another reason, in addition to the difficulty of language models to answer questions posed as an opposition or with negation, why prompt 3a performs poorly in the test datasets since the token ``against'' appears many times and in a diverse set of contexts, which makes it harder for the model to capture a useful contextual meaning (distributed representation) of it. 

Our conclusions from this ground-up approach are limited since we are unable to consider n-grams when $n>2$ due to computational expense, and it still remains to be completely understood why the performance varies across different prompts. 

\onecolumn
\section{Performance of FlanT5-XXL on SemEval Task 6A and 6B compared against baselines}~\label{app:semeval_performance}
\begin{table*}[h]
\setlength\tabcolsep{1.5pt} 
\begin{center}
\begin{tabular}{ccccccc}
\hline
Model$\backslash$Target & AT & CC & LA & FM & HC & $F_{avg}$ \\
\hline
\multirow{1}{8em}{\centering LibSVM (2017) } & \multirow{1}{4em}{\centering 80.95}
& \multirow{1}{4em}{\centering 65.1}& \multirow{1}{4em}{\centering 82.25} & \multirow{1}{4em}{\centering 79.45} & \multirow{1}{4em}{\centering 77.8} &  \multirow{1}{4em}{\centering 77.11} 
\\
\multirow{1}{8em}{\centering PE-HCN (2020)} & \multirow{1}{4em}{\centering 81.06}
& \multirow{1}{4em}{\centering 77.80} & \multirow{1}{4em}{\centering 71.76} & \multirow{1}{4em}{\centering 74.10} & \multirow{1}{4em}{\centering 72.33} &  \multirow{1}{4em}{\centering \textbf{78.43}}
\\
\multirow{1}{8em}{\centering TimeLM (2022) } & \multirow{1}{4em}{\centering -}
& \multirow{1}{4em}{\centering -}& \multirow{1}{4em}{\centering -} & 
\multirow{1}{4em}{\centering -} & 
\multirow{1}{4em}{\centering -} &  \multirow{1}{4em}{\centering 72.9} 
\\
\multirow{1}{8em}{\centering ChatGPT (2023) } & \multirow{1}{4em}{\centering -}
& \multirow{1}{4em}{\centering -}& \multirow{1}{4em}{\centering -} & 
\multirow{1}{4em}{\centering -} & 
\multirow{1}{4em}{\centering -} &  \multirow{1}{4em}{\centering 56.44} 
\\
\hline
\multirow{1}{9em}{\centering FlanT5 - 3f (2)} & \multirow{1}{4em}{\centering 72.22}
& \multirow{1}{4em}{\centering 72.88}& \multirow{1}{4em}{\centering 65.65} & \multirow{1}{4em}{\centering 64.52} & \multirow{1}{4em}{\centering \textbf{82.07}} &  \multirow{1}{4em}{\centering 75.60} 
\\
\multirow{2}{9em}{\centering FlanT5 - 3f (mean)\\ (SME)} & \multirow{2}{4em}{\centering 74.70 \\ (0.43)}
& \multirow{2}{4em}{\centering 73.08 \\ (1.09)}& \multirow{2}{4em}{\centering 67.75 \\ (0.65)} & \multirow{2}{4em}{\centering 61.37 \\ (0.76)} & \multirow{2}{4em}{\centering 79.76 \\ (0.92)} &  \multirow{2}{4em}{\centering 74.54 \\ (0.37)}
\\\\
\hline
\end{tabular}
\caption{$F_{avg}$ scores of FlanT5-XXL on Sem Eval 2016 Task 6A. 3f refers to prompt 3f, and (7) to the instruction ID 7. SME is the standard mean error across the 9 instructions.}~\label{tab:taskA}
\end{center}
\end{table*}

\begin{table*}[h]
\setlength\tabcolsep{1.5pt} 
\begin{center}
\begin{tabular}{cc}
\hline
Model$\backslash$Target & $F_{avg}$ \\
\hline
\multirow{1}{8em}{\centering BiLSTM (2016) } & \multirow{1}{4em}{\centering 58.03}
\\
\multirow{1}{8em}{\centering SVM (2017) } & \multirow{1}{4em}{\centering 61.57}
\\
\multirow{1}{8em}{\centering POLITICS (2022)} & \multirow{1}{4em}{\centering 63.54}
\\
\multirow{1}{8em}{\centering PT-HCL (2022) } & \multirow{1}{4em}{\centering 50.1}
\\
\hline
\multirow{2}{9em}{\centering\textbf{FlanT5 - 1 (6)}} & \multirow{2}{4em}{\centering \textbf{67.88}}
\\\\
\multirow{2}{8em}{\centering FlanT5-1 (mean) \\ (SME)} & \multirow{2}{4em}{\centering 67.62 \\ (0.11)}
\\\\
\multirow{2}{8em}{\centering \textit{FlanT5 - 3f (2)}} & \multirow{2}{4em}{\centering \textit{67.20}}
\\\\
\multirow{2}{8em}{\centering FlanT5-3f (mean) \\ (SME)} & \multirow{2}{4em}{\centering 65.54 \\ (0.26)}
\\\\
\hline
\end{tabular}
\caption{$F_{avg}$ scores of FlanT5-XXL on SemEval 2016 Task 6B. 1 refers to prompt 1, and (6) to instruction ID 6. 3f refers to prompt 3f and (2) to instruction ID 2. The best performer is boldfaced and the second best performer, which happens to be the best in Task 6A, is shown in italics. The mean $F_{avg}$ score and its standard mean error (SME) of the two best performers across different instructions are also shown.}~\label{tab:taskB}
\end{center}
\end{table*} 
\clearpage
\newpage
\section{Performance of FlanT5-XXL on P-Stance compared against baselines}~\label{app:pstance_performance}
\begin{table*}[h]
\setlength\tabcolsep{1.5pt} 
\begin{center}
\begin{tabular}[h]{ccccc}
\hline
Model$\backslash$Target & DT & JB & BS & $F_{avg}$ \\
\hline
\multirow{1}{14em}{\centering RoBERTa KEPT (2023) } & \multirow{1}{4em}{\centering 89.5}
& \multirow{1}{4em}{\centering 88.6}& \multirow{1}{4em}{\centering -} & \multirow{1}{4em}{\centering -} 
\\
\multirow{1}{14em}{\centering RoBERTa KSCP (2022)} & \multirow{1}{4em}{\centering 86.8}
& \multirow{1}{4em}{\centering 86.9} & \multirow{1}{4em}{\centering 80.2} & \multirow{1}{4em}{\centering -}
\\
\multirow{1}{14em}{\centering RoBERTa KSCP CT (2022)} & \multirow{1}{4em}{\centering 63.4}
& \multirow{1}{4em}{\centering 72.45} & \multirow{1}{4em}{\centering 63.95} & \multirow{1}{4em}{\centering -}
\\
\multirow{1}{14em}{\centering COT ChatGPT (2023) } & \multirow{1}{4em}{\centering 85.7}
& \multirow{1}{4em}{\centering 82.8}& \multirow{1}{4em}{\centering 80.8} & 
\multirow{1}{4em}{\centering -}
\\
\multirow{1}{14em}{\centering Zero Shot ChatGPT (2023) } & \multirow{1}{4em}{\centering 83.2}
& \multirow{1}{4em}{\centering 82.0}& \multirow{1}{4em}{\centering 79.4} & 
\multirow{1}{4em}{\centering -}  
\\
\hline
\multirow{2}{9em}{\centering FlanT5 - greedy \\ (Prompt, Inst. ID)} & \multirow{2}{6em}{\centering 84.43 \\ (3f,5)}
& \multirow{2}{6em}{\centering  84.28\\ (4f\_not,7) }& 
\multirow{2}{6em}{\centering  81.45 \\(3f,6) } & 
\multirow{2}{6em}{\centering  82.88\\(3f,4) } 
\\\\
\multirow{2}{9em}{\centering FlanT5 - PMI/AfT \\ (Prompt, Inst. ID)} & 
\multirow{2}{4em}{\centering  84.23\\ (3f,4)}
& 
\multirow{2}{4em}{\centering  84.22 \\ (4f\_not,8)}& 
\multirow{2}{4em}{\centering  82.29 \\  (3f,8)} & 
\multirow{2}{4em}{\centering  82.99\\(3f,1) }
\\\\
\hline
\end{tabular}
\caption{$F_{avg}$ scores of FlanT5-XXL on P-Stance compared to the SoTA baselines. The scores in RoBERTa KSCP CT (2022) are obtained by averaging the cross-target scores for each target shown in Fig.~\ref{fig:pstance}. 
For instance, the score for Donald Trump is obtained by averaging the scores of RoBERTa KSCP bernie (2022) and RoBERTa KSCP biden (2022). We note that FlanT5-XXL outperforms the SoTA Zero shot baseline---Zero Shot ChatGPT(2023)---in PMI, AfT, and greedy decoding. }~\label{tab:pstance}
\end{center}
\end{table*}
\clearpage
\newpage
\twocolumn 
\section{SemEval 2016 Task 6A $F_{avg}$ scores from greedy decoding}~\label{app:taskA}

\begin{table}[h]
\setlength{\tabcolsep}{3pt}
\centering
\begin{tabular}{ccccccc}
\toprule
  Inst.\textbackslash ID &     1 &     2 &    3f &    3a &  4f\_not &  4a\_not \\
\midrule
                  1 & 60.81 & 68.78 & 74.31 & 66.69 &   65.24 &   42.79 \\
                  2 & 59.45 & 71.94 & 72.22 & 59.36 &   66.59 &   42.12 \\
                  3 & 62.67 & 72.38 & 73.64 & 59.69 &   67.33 &   43.16 \\
                  4 & 62.15 & 71.28 & 75.60 & 62.09 &   64.33 &   43.30 \\
                  5 & 62.50 & 73.93 & 76.18 & 68.08 &   66.35 &   39.63 \\
                  6 & 62.05 & 71.32 & 73.86 & 71.99 &   67.04 &   43.38 \\
                  7 & 58.53 & 72.20 & 75.77 & 66.97 &   64.69 &   42.25 \\
                  8 & 61.85 & 72.51 & 75.29 & 62.89 &   65.27 &   46.15 \\
                 9 & 62.93 & 72.03 & 75.47 & 62.09 &   64.78 &   39.95 \\
\bottomrule
\end{tabular}
\caption{$F_{avg}$ scores for Atheism}
\end{table}


\begin{table}[h]
\setlength{\tabcolsep}{3pt}
\centering
\begin{tabular}{ccccccc}
\toprule
 Inst.\textbackslash ID &     1 &     2 &    3f &    3a &  4f\_not &  4a\_not \\
\midrule
                  1 & 65.14 & 56.67 & 62.05 & 56.22 &   58.95 &   59.54 \\
                  2 & 65.72 & 59.88 & 64.52 & 52.36 &   57.31 &   59.88 \\
                  3 & 65.88 & 56.31 & 62.71 & 55.25 &   57.07 &   59.73 \\
                  4 & 64.79 & 56.44 & 62.66 & 55.78 &   57.82 &   58.00 \\
                  5 & 64.47 & 57.75 & 60.11 & 59.24 &   60.35 &   57.20 \\
                  6 & 65.38 & 55.10 & 56.89 & 56.82 &   60.21 &   59.25 \\
                  7 & 65.19 & 54.28 & 59.25 & 58.95 &   58.95 &   57.39 \\
                  8 & 65.23 & 58.51 & 61.78 & 55.91 &   57.07 &   57.40 \\
                 9 & 65.25 & 54.14 & 62.34 & 55.58 &   56.88 &   58.30 \\
\bottomrule
\end{tabular}
\caption{$F_{avg}$ scores for Feminist Movement}
\end{table}

\begin{flushright}
\begin{table}[h]
\setlength{\tabcolsep}{3pt}
\
\begin{tabular}{ccccccc}
\toprule
  Inst.\textbackslash ID &     1 &     2 &    3f &    3a &  4f\_not &  4a\_not \\
\midrule
                  1 & 61.62 & 71.69 & 70.28 & 75.89 &   64.74 &   63.64 \\
                  2 & 58.17 & 73.26 & 72.88 & 75.89 &   65.09 &   59.52 \\
                  3 & 54.19 & 69.69 & 75.43 & 75.89 &   62.16 &   60.99 \\
                  4 & 57.37 & 75.62 & 73.30 & 74.17 &   63.80 &   64.58 \\
                  5 & 60.29 & 71.15 & 77.37 & 71.10 &   62.19 &   65.77 \\
                  6 & 53.46 & 71.63 & 76.32 & 69.88 &   60.75 &   64.42 \\
                  7 & 59.93 & 76.43 & 68.38 & 66.82 &   61.83 &   62.80 \\
                  8 & 56.46 & 72.74 & 74.88 & 74.17 &   64.86 &   62.53 \\
                 9 & 66.02 & 76.59 & 68.89 & 72.58 &   66.28 &   64.03 \\
\bottomrule
\end{tabular}
\caption{$F_{avg}$ scores for Climate Change is a Real Concern}
\end{table}
\end{flushright}


\begin{table}[h]
\setlength{\tabcolsep}{3pt}
\centering
\begin{tabular}{ccccccc}
\toprule
 Inst.\textbackslash ID &     1 &     2 &    3f &    3a &  4f\_not &  4a\_not \\
\midrule
                  1 & 79.81 & 77.04 & 82.19 & 63.01 &   70.55 &   65.94 \\
                  2 & 79.47 & 77.69 & 82.07 & 61.39 &   66.93 &   67.15 \\
                  3 & 81.30 & 78.98 & 81.70 & 65.47 &   69.07 &   66.32 \\
                  4 & 80.76 & 74.98 & 79.50 & 64.57 &   70.34 &   63.96 \\
                  5 & 79.00 & 79.18 & 80.86 & 68.17 &   72.33 &   64.20 \\
                  6 & 79.44 & 77.90 & 81.58 & 74.14 &   75.41 &   63.92 \\
                  7 & 78.48 & 75.36 & 73.77 & 66.30 &   71.40 &   63.42 \\
                  8 & 78.18 & 76.29 & 77.80 & 66.25 &   71.18 &   65.54 \\
                 9 & 80.03 & 76.78 & 78.33 & 65.67 &   70.03 &   64.74 \\
\bottomrule
\end{tabular}
\caption{$F_{avg}$ scores for Hillary Clinton}
\end{table}


\begin{flushright}
\begin{table}[h]
\setlength{\tabcolsep}{3pt}
\centering
\begin{tabular}{ccccccc}
\toprule
  Inst.\textbackslash ID &     1 &     2 &    3f &    3a &  4f\_not &  4a\_not \\
\midrule
                  1 & 70.25 & 65.80 & 64.11 & 60.62 &   64.19 &   54.53 \\
                  2 & 69.54 & 66.17 & 65.65 & 55.82 &   61.94 &   57.77 \\
                  3 & 70.43 & 66.54 & 68.50 & 61.27 &   63.15 &   55.14 \\
                  4 & 69.35 & 65.39 & 69.43 & 62.58 &   64.39 &   55.48 \\
                  5 & 69.98 & 69.25 & 67.47 & 63.00 &   66.42 &   53.08 \\
                  6 & 70.35 & 65.71 & 66.60 & 65.97 &   66.18 &   52.99 \\
                  7 & 68.16 & 64.86 & 69.64 & 60.07 &   64.79 &   54.28 \\
                  8 & 69.52 & 65.91 & 69.04 & 62.01 &   64.39 &   54.32 \\
                 9 & 67.82 & 64.77 & 69.34 & 59.82 &   64.03 &   55.66 \\
\bottomrule
\end{tabular}
\caption{$F_{avg}$ scores for Legalization of Abortion}
\end{table}
\end{flushright}

\clearpage
\newpage
\section{SemEval 2016 Task 6B $F_{avg}$ scores from greedy decoding}~\label{app:taskB}
\begin{table}[h]
\setlength{\tabcolsep}{3pt}
\centering
\begin{tabular}{ccccccc}
\toprule
 Inst.\textbackslash ID &     1 &     2 &    3f &    3a &  4f\_not &  4a\_not \\
\midrule
                  1 & 67.74 & 61.15 & 65.54 & 54.26 &   59.67 &   60.94 \\
                  2 & 67.01 & 62.49 & 67.20 & 51.70 &   57.47 &   61.48 \\
                  3 & 68.13 & 62.90 & 65.91 & 55.63 &   59.92 &   60.81 \\
                  4 & 67.78 & 60.66 & 65.46 & 53.12 &   59.16 &   59.39 \\
                  5 & 67.33 & 63.28 & 65.26 & 55.84 &   61.19 &   57.06 \\
                  6 & 67.88 & 62.58 & 65.55 & 60.33 &   60.77 &   59.39 \\
                  7 & 67.66 & 60.57 & 64.19 & 53.12 &   59.60 &   59.08 \\
                  8 & 67.47 & 60.69 & 65.39 & 53.62 &   59.68 &   60.01 \\
                 9 & 67.56 & 62.08 & 65.39 & 52.81 &   60.04 &   59.41 \\
\bottomrule
\end{tabular}
\caption{$F_{avg}$ scores for Donald Trump}
\end{table}
\clearpage
\newpage
\section{P-Stance $F_{avg}$ scores using greedy decoding}~\label{app:pstance_greedy}
\begin{table}[h]
\setlength{\tabcolsep}{3pt}
\centering
\begin{tabular}{ccccccc}
\toprule
 Inst.\textbackslash ID &     1 &     2 &    3f &    3a &  4f\_not &  4a\_not \\
\midrule
                  1 & 74.07 & 71.87 & 80.46 & 72.37 &   81.80 &   62.39 \\
                  2 & 73.50 & 81.10 & 78.70 & 78.14 &   81.54 &   61.29 \\
                  3 & 73.18 & 71.54 & 81.14 & 77.78 &   80.01 &   63.44 \\
                  4 & 73.61 & 83.03 & 80.53 & 79.11 &   82.93 &   62.30 \\
                  5 & 74.59 & 80.46 & 84.43 & 72.37 &   79.91 &   64.48 \\
                  6 & 73.01 & 76.04 & 84.04 & 72.99 &   79.79 &   64.48 \\
                  7 & 72.68 & 82.54 & 84.16 & 76.85 &   82.13 &   62.39 \\
                  8 & 72.47 & 82.58 & 81.96 & 77.17 &   81.92 &   63.55 \\
                 9 & 73.70 & 82.22 & 79.81 & 79.11 &   82.23 &   61.90 \\
\bottomrule
\end{tabular}
\caption{$F_{avg}$ scores for Donald Trump}
\end{table}

\begin{table}[h]
\setlength{\tabcolsep}{3pt}
\centering
\begin{tabular}{ccccccc}
\toprule
 Inst.\textbackslash ID &     1 &     2 &    3f &    3a &  4f\_not &  4a\_not \\
\midrule
                  1 & 82.40 & 62.85 & 73.35 & 77.96 &   83.87 &   79.41 \\
                  2 & 81.68 & 77.17 & 75.68 & 81.52 &   84.18 &   76.93 \\
                  3 & 82.51 & 68.73 & 77.02 & 83.15 &   83.50 &   80.05 \\
                  4 & 82.50 & 80.24 & 77.97 & 82.99 &   83.05 &   79.02 \\
                  5 & 82.55 & 78.71 & 82.73 & 81.85 &   84.23 &   81.53 \\
                  6 & 82.24 & 72.32 & 81.17 & 81.72 &   83.66 &   81.24 \\
                  7 & 82.20 & 78.79 & 79.83 & 83.35 &   84.28 &   79.29 \\
                  8 & 82.63 & 79.06 & 79.26 & 82.69 &   83.67 &   79.57 \\
                 9 & 82.24 & 77.58 & 78.48 & 83.09 &   83.44 &   79.23 \\
\bottomrule
\end{tabular}
\caption{$F_{avg}$ scores for Joe Biden}
\end{table}

\newpage

\begin{table}[h]
\setlength{\tabcolsep}{3pt}
\centering
\begin{tabular}{ccccccc}
\toprule
 Inst.\textbackslash ID &     1 &     2 &    3f &    3a &  4f\_not &  4a\_not \\
\midrule
                  1 & 77.77 & 75.52 & 77.78 & 75.32 &   79.51 &   69.60 \\
                  2 & 77.29 & 79.97 & 78.72 & 77.22 &   80.41 &   66.58 \\
                  3 & 76.96 & 77.12 & 80.79 & 77.62 &   80.00 &   71.55 \\
                  4 & 76.80 & 79.09 & 79.53 & 78.47 &   80.98 &   69.48 \\
                  5 & 78.09 & 78.44 & 80.61 & 77.94 &   80.31 &   71.42 \\
                  6 & 76.48 & 78.19 & 81.45 & 78.10 &   80.16 &   71.44 \\
                  7 & 77.28 & 78.64 & 80.60 & 79.04 &   80.60 &   68.54 \\
                  8 & 76.00 & 79.36 & 80.94 & 78.82 &   80.88 &   69.72 \\
                 9 & 77.44 & 79.17 & 80.47 & 78.79 &   81.13 &   69.33 \\
\bottomrule
\end{tabular}
\caption{$F_{avg}$ scores for Bernie Sanders}
\end{table}

\clearpage
\newpage
\section{P-Stance $F_{avg}$ scores using PMI decoding}~\label{app:pstance_pmi}
\begin{table}[h]
\setlength{\tabcolsep}{3pt}
\centering
\begin{tabular}{ccccccc}
\toprule
 Inst.\textbackslash ID &     1 &     2 &    3f &    3a &  4f\_not &  4a\_not \\
\midrule
                  1 & 80.54 & 76.98 & 83.34 & 78.02 &   82.55 &   62.28 \\
                  2 & 78.66 & 81.60 & 81.97 & 77.66 &   81.61 &   68.79 \\
                  3 & 80.11 & 75.14 & 82.33 & 78.08 &   81.26 &   67.04 \\
                  4 & 80.88 & 82.39 & 84.23 & 81.26 &   82.40 &   68.96 \\
                  5 & 79.27 & 80.50 & 81.97 & 77.10 &   81.97 &   68.33 \\
                  6 & 80.54 & 80.43 & 82.61 & 77.23 &   81.73 &   69.35 \\
                  7 & 79.44 & 81.26 & 81.33 & 78.66 &   82.34 &   70.38 \\
                  8 & 79.33 & 82.10 & 81.70 & 78.54 &   82.76 &   69.57 \\
                 9 & 81.63 & 78.54 & 83.47 & 80.84 &   82.16 &   68.73 \\
\bottomrule
\end{tabular}
\caption{$F_{avg}$ scores for Donald Trump}
\end{table}


\begin{table}[ht]
\setlength{\tabcolsep}{3pt}
\centering
\begin{tabular}{ccccccc}
\toprule
 Inst.\textbackslash ID &     1 &     2 &    3f &    3a &  4f\_not &  4a\_not \\
\midrule
                  1 & 80.87 & 79.67 & 83.67 & 83.09 &   83.31 &   79.70 \\
                  2 & 80.19 & 80.60 & 81.77 & 82.12 &   83.43 &   81.14 \\
                  3 & 81.92 & 77.29 & 81.55 & 82.47 &   84.01 &   81.90 \\
                  4 & 81.19 & 83.51 & 81.80 & 83.44 &   83.90 &   81.04 \\
                  5 & 82.00 & 81.38 & 82.54 & 82.45 &   83.27 &   81.75 \\
                  6 & 81.18 & 81.66 & 80.86 & 82.14 &   83.35 &   81.73 \\
                  7 & 80.93 & 81.10 & 80.52 & 81.67 &   83.70 &   80.75 \\
                  8 & 82.00 & 83.16 & 81.50 & 81.65 &   84.22 &   81.16 \\
                 9 & 82.79 & 83.72 & 81.77 & 82.77 &   83.50 &   81.46 \\
\bottomrule
\end{tabular}
\caption{$F_{avg}$ scores for Joe Biden}
\end{table}

\begin{table}[ht]
\setlength{\tabcolsep}{3pt}
\centering
\begin{tabular}{ccccccc}
\toprule
 Inst.\textbackslash ID &     1 &     2 &    3f &    3a &  4f\_not &  4a\_not \\
\midrule
                  1 & 75.45 & 72.89 & 81.53 & 78.43 &   79.98 &   69.30 \\
                  2 & 75.43 & 74.64 & 81.49 & 77.76 &   79.81 &   70.17 \\
                  3 & 76.44 & 77.42 & 81.35 & 77.12 &   79.37 &   71.42 \\
                  4 & 74.55 & 78.39 & 81.54 & 77.76 &   80.46 &   70.54 \\
                  5 & 75.43 & 76.31 & 80.28 & 77.76 &   80.14 &   70.54 \\
                  6 & 75.07 & 76.95 & 80.55 & 76.95 &   79.99 &   70.57 \\
                  7 & 73.56 & 77.91 & 80.38 & 76.77 &   79.52 &   70.91 \\
                  8 & 75.74 & 80.63 & 82.29 & 73.63 &   80.31 &   72.91 \\
                 9 & 75.92 & 78.70 & 81.52 & 78.39 &   79.37 &   71.38 \\
\bottomrule
\end{tabular}
\caption{$F_{avg}$ scores for Bernie Sanders}
\end{table}

\clearpage
\newpage
\section{P-Stance $F_{avg}$ scores using AfT decoding}~\label{app:pstance_aft}
\begin{table}[h]
\setlength{\tabcolsep}{3pt}
\centering
\begin{tabular}{ccccccc}
\toprule
 Inst.\textbackslash ID &     1 &     2 &    3f &    3a &  4f\_not &  4a\_not \\
\midrule
                  1 & 80.54 & 76.98 & 81.95 & 78.02 &   82.55 &   62.28 \\
                  2 & 78.66 & 81.60 & 47.24 & 77.66 &   81.61 &   65.10 \\
                  3 & 80.11 & 75.14 & 82.33 & 78.08 &   81.26 &   67.04 \\
                  4 & 80.88 & 82.39 & 84.23 & 81.26 &   82.40 &   68.96 \\
                  5 & 79.27 & 80.50 & 81.72 & 77.10 &   81.97 &   68.33 \\
                  6 & 80.54 & 80.43 & 82.10 & 77.23 &   81.73 &   69.35 \\
                  7 & 79.44 & 81.26 & 81.33 & 78.66 &   82.34 &   70.38 \\
                  8 & 79.33 & 82.10 & 81.70 & 78.54 &   82.76 &   69.57 \\
                  9 & 81.63 & 78.54 & 83.47 & 80.84 &   82.16 &   68.73 \\
\bottomrule
\end{tabular}
\caption{$F_{avg}$ scores for Donald Trump}
\end{table}
\begin{table}[h]
\setlength{\tabcolsep}{3pt}
\centering
\begin{tabular}{ccccccc}
\toprule
 Inst.\textbackslash ID &     1 &     2 &    3f &    3a &  4f\_not &  4a\_not \\
\midrule
                  1 & 80.87 & 79.67 & 79.43 & 83.09 &   83.31 &   79.70 \\
                  2 & 80.19 & 80.60 & 39.36 & 82.12 &   83.43 &   81.14 \\
                  3 & 81.92 & 77.29 & 81.55 & 82.47 &   84.01 &   81.90 \\
                  4 & 81.19 & 83.51 & 81.80 & 83.44 &   83.90 &   81.04 \\
                  5 & 82.00 & 81.38 & 81.77 & 82.45 &   83.27 &   81.77 \\
                  6 & 81.18 & 81.66 & 79.30 & 82.14 &   83.35 &   81.48 \\
                  7 & 80.93 & 81.10 & 80.52 & 81.67 &   83.70 &   80.75 \\
                  8 & 82.00 & 83.16 & 81.50 & 81.65 &   84.22 &   81.16 \\
                  9 & 82.79 & 83.72 & 81.77 & 82.77 &   83.50 &   81.46 \\
\bottomrule
\end{tabular}
\caption{$F_{avg}$ scores for Joe Biden}
\end{table}
\begin{table}[h]
\setlength{\tabcolsep}{3pt}
\centering
\begin{tabular}{ccccccc}
\toprule
 Inst.\textbackslash ID &     1 &     2 &    3f &    3a &  4f\_not &  4a\_not \\
\midrule
                  1 & 75.45 & 72.89 & 81.53 & 78.43 &   79.98 &   69.30 \\
                  2 & 75.43 & 74.64 & 81.49 & 77.76 &   79.81 &   70.17 \\
                  3 & 76.44 & 77.42 & 81.35 & 77.12 &   79.37 &   71.42 \\
                  4 & 74.55 & 78.39 & 81.54 & 77.76 &   80.46 &   70.54 \\
                  5 & 75.43 & 76.31 & 80.28 & 77.92 &   80.14 &   70.54 \\
                  6 & 75.07 & 76.95 & 80.55 & 76.95 &   79.99 &   70.57 \\
                  7 & 73.56 & 77.91 & 80.38 & 76.77 &   79.52 &   70.91 \\
                  8 & 75.74 & 80.63 & 82.29 & 73.63 &   80.31 &   72.91 \\
                  9 & 75.92 & 78.70 & 81.52 & 78.39 &   79.37 &   71.38 \\
\bottomrule
\end{tabular}
\caption{$F_{avg}$ scores for Bernie Sanders}
\end{table}
\clearpage
\newpage
\section{Statistical significance}\label{app:statsig}

\begin{figure}[ht]
    \includegraphics[width=0.45\textwidth, trim={0 0in 0 0in}]{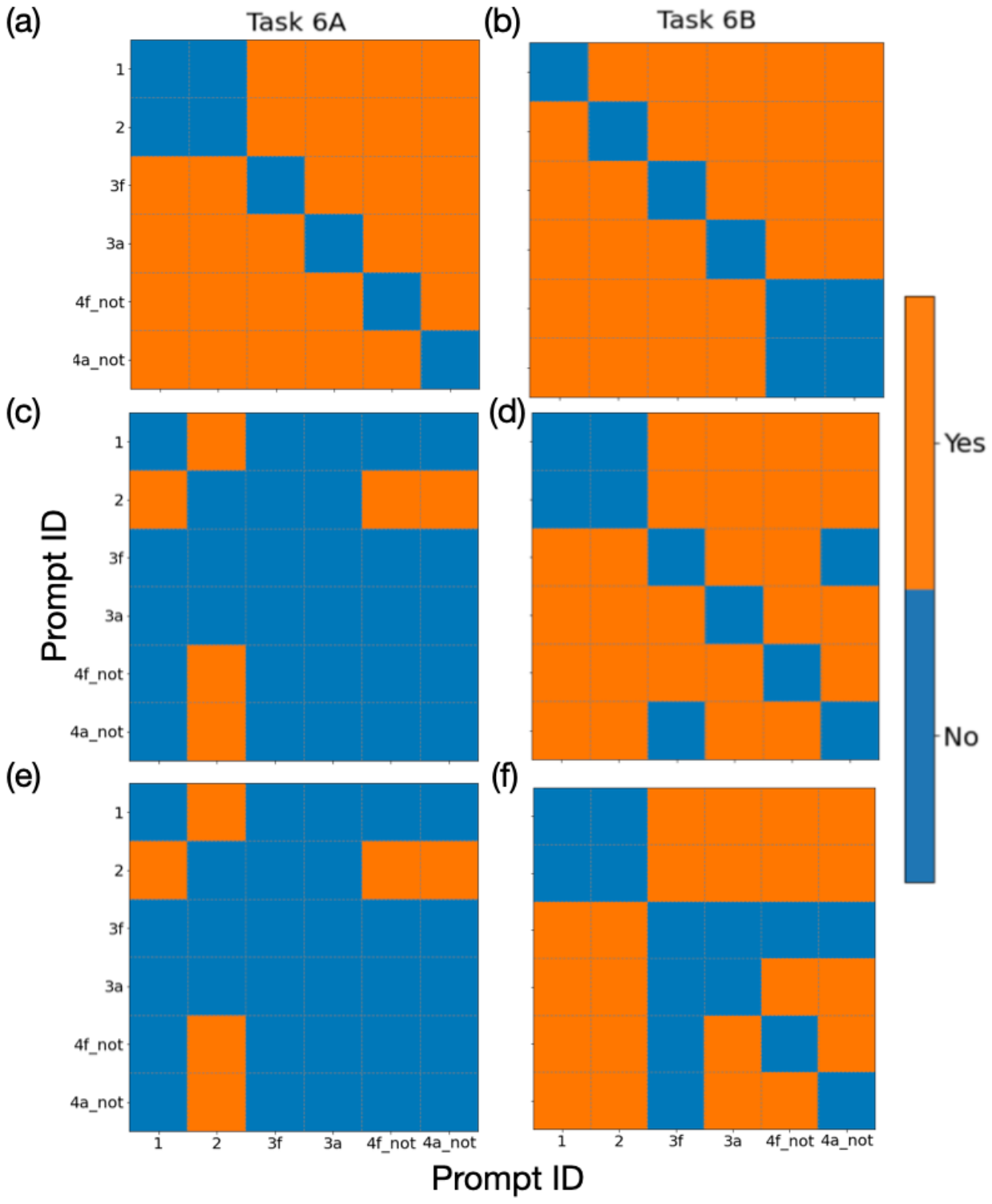}
    \caption{Statistical significance of the performance, measured via $F_{avg}$ scores, between different prompts on the SemEval 2016 Task 6 dataset. The top row is greedy decoding, the middle row is PMI decoding, and the bottom row is AfT decoding. ``Yes'' and ``No'' indicates p-value $<0.05$ and $p>0.05$, respectively, obtained independent sample t-test.}
    \label{fig:statsig_semeval}
\end{figure}

An independent sample t-test showed that the difference in the mean $F_{avg}$ scores between prompt is statistically significant ($p<0.05$), except in SemEval Task 6A---pair (1,2) from greedy decoding. 
In PMI, and AFT decoding, just pairs (1,2), (2,4f\_not), and (2,4a\_not) have statistically significant differences. 
In AfT decoding, pairs (1,2), (3f,3a
In SemEval Task 6B---pair (4f\_not,4a\_not) from greedy decoding, and pairs (1,2), (3f,4a\_not) from PMI decoding do not have statistically significant differences.
In AfT decoding, pairs (1,2), (3f,3a), (3f,4f\_not), (3f,4a\_not) do not have statistically significant differences. 

\begin{figure}[ht]
    \hspace*{-1cm}
    \includegraphics[width=0.6\textwidth, trim={0 0in 0 0in}]{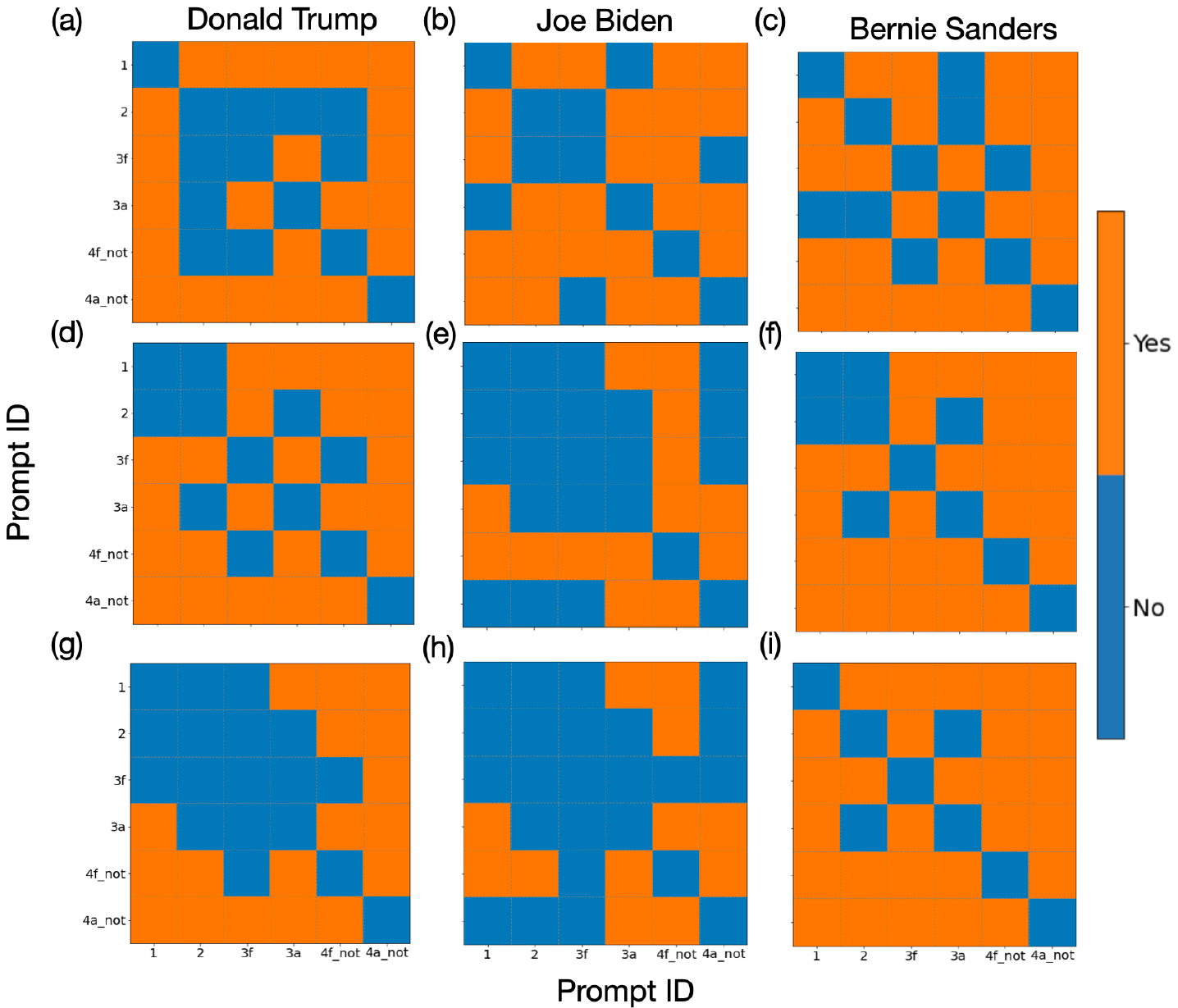}
    \caption{Statistical significance of the performance, measured via $F_{avg}$ scores, between different prompts on the P-Stance dataset. The top row is greedy decoding, the middle row is PMI decoding, and the bottom row is AfT decoding. ``Yes'' and ``No'' indicates p-value $<0.05$ and $p>0.05$, respectively, obtained independent sample t-test.}
    \label{fig:statsig_pstance}
\end{figure}

An independent sample t-test was performed to gauge the statistical significance of the difference in the mean $F_{avg}$ scores between prompt pairs. The differences are mostly statistically significant ($p<0.05$) for greedy decoding except in a few cases---for Donald Trump pairs (2,3f), (2,3a), (2,4f\_not), (3f,4f\_not), for Joe Biden pairs (1,3a), (2,3f), (3f,4a\_not), and for Bernie Sanders pairs (1,3a), (2,3a), (3f,4f\_not). 
There exists statistically significant differences ($p<0.05$) for PMI decoding except in cases---for Donald Trump pairs (1,2), (2,3a), (3f,4f\_not), for Joe Biden pairs (1,2), (1,3f), (1,4f\_not), (3f,2), (3a,2), (3a,3f), (2,4a\_not), (3f, 4a\_not), and for Bernie Sanders pairs (1,2), (3a,2). 
In case of AfT decoding, significant differences ($p<0.05$) exist except in cases---(2,1),(3f,1),(3f,2),(3a,2),(3a,3f),(4f\_not,3f) for Donald Trump, 
(2,1),(3f,1),(3f,2),(3a,2),(3a,3f), (4f\_not,3f), (4a\_not,1), (4a\_not,2), (4a\_not,3f) for Joe Biden, and (2,3a) for Bernie Sanders. 
Overall, the difference between prompt pairs is almost always significant ($p<0.05$)for Bernie Sanders regardless of decoding strategy, is mostly significant when greedy decoding is used for Joe Biden, and is significant when greedy or PMI decoding is used for Donald Trump. 

\clearpage
\newpage
\twocolumn
\section{Contamination of training data}~\label{app:contamination}
LLMs are pre-trained on a large corpus of web-text collected by the Common Crawl Project which produces about 20TB of text data extracted from crawling web pages each month.\footnote{https://commoncrawl.org/}
Like all machine learning systems, the quality of data impacts their performance, and it is essential to ensure that the test data on which these LLMs are evaluated on, is not present in its pre-training or fine-tuning dataset~\citep{aiyappa-etal-2023-trust}. 
While many frameworks to document the properties of training data have been proposed~\citep{gebru2021datasheets}, proper documentation of such a large corpus of web text is still in its infancy~\citep{dodge2021documenting}. 
Here, we wish to check if the SemEval 2016 Task 6 and the P-Stance train/validation/test datasets are a part of the pretaining and instruction finetuning data of FlanT5-XXL. If this is true, then the performance we are observing might just be a result of the model having memorized the dataset and is thus not a true evaluation of its zero-shot stance detection capability. 

\subsection{Pre-training data: C4}~\label{app:C4}
T5 (and thus FlanT5) is pre-trained on a filtered (cleaned) version of Common Crawl data from April 2019 called ``Colossal Clean Crawled Corpus (C4)~\citep{raffel2020exploring}.'' Recent work identified evaluation examples from various NLP benchmark datasets to be a part of C4 thus questioning the claimed capabilities of models pre-trained on C4~\citep{dodge2021documenting}. Since the P-Stance dataset was collected during the 2020 presidential elections from Twitter, and the C4 used to pre-train T5 is from before (i.e., April 2019), we argue that P-Stance dataset cannot be present in the pertaining data. 

In order to try and ensure that the SemEval 2016 Task 6 dataset is not present in C4, we first checked whether the tweets present in the test sets of tasks A and B, and the train set of task A are present in verbatim or in lowercase as individual lines in the C4 dataset hosted by Hugging Face\footnote{https://huggingface.co/datasets/allenai/c4/tree/main/en} and found no evidence of this. 
%

Next, we used the indexed C4 GUI developed by AllenAI\footnote{https://c4-search.apps.allenai.org/} to search for the potential existence of the SemEval 2016 Task 6 data in C4. Although we did notice that the website hosting the call for the submission to competition was indexed in the dataset\footnote{http://bit.ly/3yMq3l5}, we did not find the actual dataset itself and thus believe it is not present in an easily recognizable manner. We also checked for the presence of random tweets from the dataset and found none of them. 

As a final step, we searched for the hashtag \#SemST in the indexed dataset\footnote{http://bit.ly/3yMq3l5} as this hashtag, for unclear reasons, was added to all the tweets in the SemEval 2016 Task 6 dataset. We did not find this hashtag in the search. 

These three findings, though not complete proof, make it likely that the SemEval 2016 Task 6 (and trivially, the P-Stance) dataset is not present in the pertaining corpus.

\subsubsection{Instruction Tuning Data}~\label{sec:instruction-tuning}

FlanT5 takes the T5 model and instruction tunes on a number of NLP datasets in a few-shot, zero-shot, or chain of thought  setting using instruction tuning methods like input inversion, template generation, etc~\citep{flan}. We performed a qualitative inspection of these datasets~\citep{chung2022scaling} and found no evidence of the SemEval 2016 Task 6 or the P-Stance dataset being present. However, we found that the model has been instruction-tuned on the Natural Instruction V2 dataset~\citep{wang2022super}---a collection of 1600+ NLP tasks with instructions---which contains within it stance detection tasks. 
For example, task 513\footnote{http://bit.ly/3YWaUs6}, with the definition ``You will be given a topic and an argument. Decide the argument's stance towards that topic. The argument's stance is in favor or against the topic. If the argument supports that topic, answer with `in favor'; otherwise, if the argument opposes the topic, answer with `against'.'' has 7 mentions of ``abortion'' and 1 mention of ``climate change.'' The input text in task 513 does not consist of tweets but is sourced from Debatepedia~\citep{kobbe2020unsupervised}.
Similarly, task 209\footnote{http://bit.ly/3FvYMac} appears to be the duplicate of 513 with the same data source and data, but with a different instruction ``Given the Target and Argument texts detect the stance that the argument has towards the topic. There are three types of stances `in favor', `against', and `neutral'.'' 
Both tasks 209 and 513 are balanced in terms of class labels~\citep{kobbe2020unsupervised}. 
The collection also has a stance detection dataset in Spanish and Catalan~\citep{zotova2020multilingual}---Task 1646\footnote{https://bit.ly/3llyuRp}---with the definition ``In this task, we have Spanish and Catalan tweets for automatic stance detection. The data has three labels Against, Favor, and Neutral which express the stance toward the target---independence of Catalonia. If the tweet criticizes the independence of Catalonia then it's `Against' and if the tweet supports it then it will be labeled as `Favor' also if the tweets state information or news rather than stating opinion then it will be characterized as `Neutral'.'' The dataset is balanced in terms of Favor and Against classes but has fewer Neutral class tweets. 
Lastly, we find task 890,\footnote{http://bit.ly/3yOL8LM} related to the stance towards global warming with the definition ``Read the passage and find if the passage agrees, disagrees, or has a neutral stance on whether Global warming is caused by human activities. Answer only with keyword (a) agrees - if passage agrees with the target (b) disagrees - if passage disagrees with the target (c) neutral - if the given passage neither agrees nor disagrees with the target. You don't need to use external knowledge in this task, and you have to answer based on the given passage.'' where the passages are news articles, published from Jan. 1, 2000 to April 12, 2020 by 63 U.S. news sources~\citep{luo2020detecting}.\footnote{Class imbalance: neutral: 873, agree: 777, disagree: 400} 

In sum, we find no evidence of SemEval 2016 Task 6 or P-Stance data in the pre-training and instruction-tuning datasets. Moreover, though FlanT5 has been instruction-tuned on certain stance detection datasets, none of these contain English tweets, which makes the stance detection setting in this work, truly zero-shot.  

\section{Improving performance}~\label{app:preprocessing}
\begin{figure*}[t]
    \centering
    \includegraphics[width=1\textwidth, trim={0 2.2in 0 2in}]{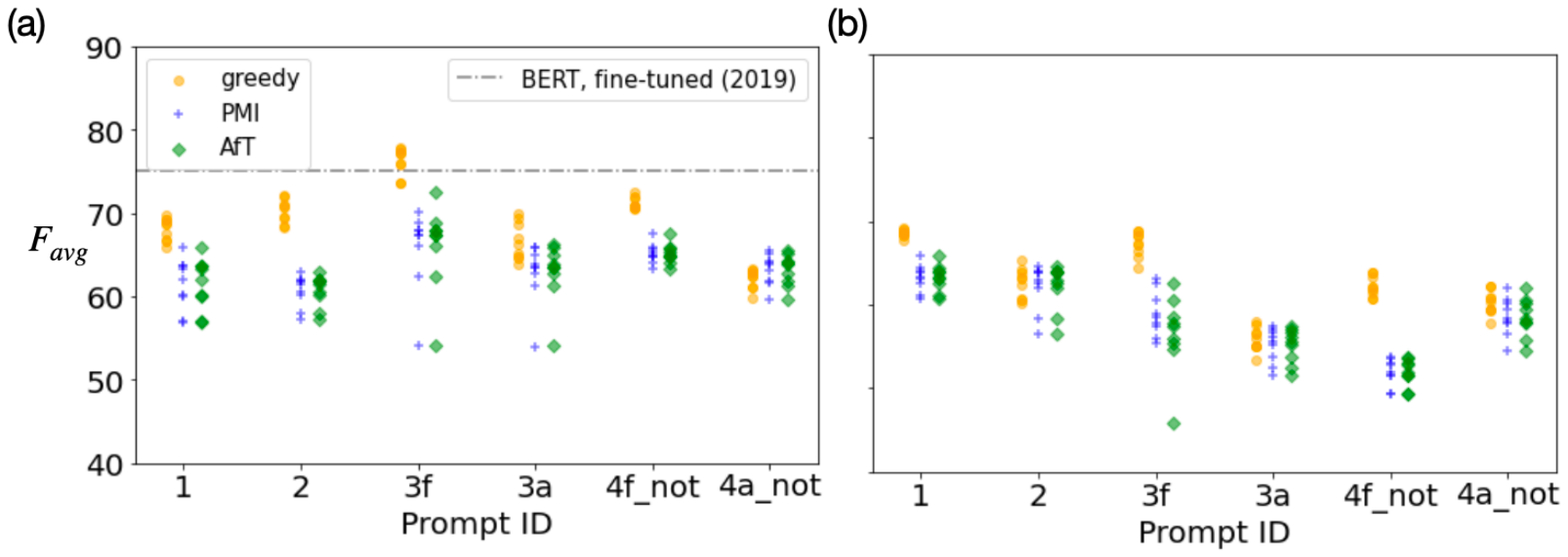}
    \caption{The $F_{avg}$ score of FlanT5-XXL on preprocessed (a) Task 6A, and (b) Task 6B of SemEval 2016 is shown in comparison against the work which proposed the pre-processing strategy. Each label on the x-axis corresponds to a prompt (see Tab.~\ref{tab:prompts}) and each point on a given prompt ID corresponds to an instruction (see Appendix.~\ref{app:instrutions}). The results of two decoding strategies---greedy and PMI---are also shown.
    }\label{fig:taskAB_preprocessed}
\end{figure*}
In addition to standard pre-processing steps such as case-folding, \citet{ghosh2019stance} expand abbreviations and split hashtags in tweets and claim that this leads to their model (finetuned BERT) performing better on the task on the SemEval 2016 Task 6. 

We adopt the same pre-processing strategy and test the performance of FlanT5-XXL on SemEval 2016 Task 6, and P-Stance dataset. 
The results on the SemEval 2016 Task 6 dataset are shown in Fig.~\ref{fig:taskAB_preprocessed} along with the the baseline model (BERT, fine-tuned (2019) from the work that proposed the pre-processing strategy \citep{ghosh2019stance}.
We note that one of our prompts, 3f (every instruction except 1, and 6), outperforms the state-of-the-art work which proposed and used the pre-processing strategy.
We also show the results of the best-performing prompt-instruction pairs (on the basis of $F_{avg}$ score) on SemEval 2016 Task 6A with pre-processing in Tab.~\ref{tab:taskA_preprocessed}.

\citet{ghosh2019stance} did not test their model's performance on Task 6B, but we have shown FlanT5-XXL's performance with pre-processing in Fig.~\ref{fig:taskAB_preprocessed}(b). 

\begin{table*}[t]
\setlength\tabcolsep{1.5pt} 
\begin{center}
\begin{tabular}[t]{ccccccc}
\hline
Model$\backslash$Target & AT & CC & LA & FM & HC & $F_{avg}$ \\
\hline
\multirow{1}{9em}{\centering FlanT5 - 3f (2)} & \multirow{1}{4em}{\centering 72.22}
& \multirow{1}{4em}{\centering 72.88}& \multirow{1}{4em}{\centering 65.65} & \multirow{1}{4em}{\centering 64.52} & \multirow{1}{4em}{\centering \textbf{82.07}} &  \multirow{1}{4em}{\centering 75.60} 
\\
\multirow{1}{9em}{\centering BERT (2019)} & \multirow{1}{4em}{\centering 74.3}
& \multirow{1}{4em}{\centering 44.6}& \multirow{1}{4em}{\centering 65.7} & \multirow{1}{4em}{\centering 65.0} & \multirow{1}{4em}{\centering 71.3} &  \multirow{1}{4em}{\centering 75.1} 
\\
\multirow{1}{9em}{\centering FlanT5-P - 3f (8)} & 
\multirow{1}{4em}{\centering \textbf{74.46} }
& \multirow{1}{4em}{\centering \textbf{77.00} }& \multirow{1}{4em}{\centering \textbf{66.31}} & \multirow{1}{4em}{\centering \textbf{71.51} } & \multirow{1}{4em}{\centering 81.01} &  \multirow{1}{4em}{\centering \textbf{77.92} }
\\\\
\end{tabular}
\caption{$F_{avg}$ scores of best performing FlanT5-XXL on SemEval 2016 Task 6A with and without preprocessing of tweets compared to the work (BERT, 2019) which proposed the preprocessing strategy~\citep{ghosh2019stance}. 3f refers to prompt 3f, and (2), (8) to the instruction ID 2, and 8 respectively. SME is the standard mean error.}~\label{tab:taskA_preprocessed}
\end{center}
\end{table*}

\begin{figure*}[t]
    \centering
    \includegraphics[width=1\textwidth, trim={0 2.5in 0 2in}]{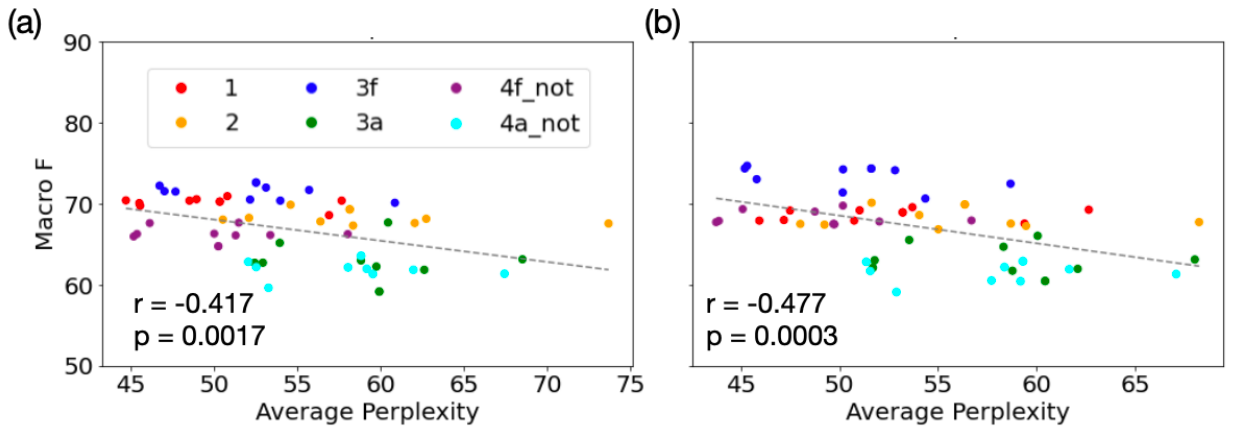}
    \caption{Correlation between prompt perplexity, per prompt ID, per instruction and $F_{avg}$ scores (from greedy) (a) before pre-processing of tweets, and (b) after pre-processing of tweets, in Task A+B. The Pearson correlation coefficient is indicated by $r$ and p-value by $p$.}\label{fig:perplexities_processing}
\end{figure*}

In Fig.~\ref{fig:perplexities_processing}, we show the average perplexities of prompts across targets, per prompt ID, per instruction ID  (Tab.~\ref{tab:prompts}) combining both the SemEval 2016 tasks, without and with the pre-processing of tweets (Fig.~\ref{fig:perplexities_processing} a and b, respectively). We note that the average perplexity with pre-processing (54.02) is less than without pre-processing (54.68), but this difference is not significant based on an independent sample t-test. We also observe a stronger correlation coefficient (-0.49) with pre-processing than without (-0.423). 

We did not notice a significant change in performance when the P-Stance dataset was preprocessed similarly.

\end{document}